\documentclass[10pt,twocolumn,letterpaper]{article} 

\usepackage{cvpr}              

\usepackage[dvipsnames]{xcolor}

\usepackage{graphicx}
\usepackage{tikz}
\usepackage{comment}
\usepackage{dirtytalk}
\usepackage{amsmath,amssymb} 
\usepackage{color}
\usepackage{graphicx}
\usepackage{amsmath}
\usepackage{amssymb}
\usepackage{booktabs}
\usepackage{times}
\usepackage{epsfig}
\usepackage{amsfonts}
\usepackage{dsfont}
\usepackage{multirow}
\usepackage{enumitem}
\usepackage{mathtools}
\usepackage{floatrow}
\usepackage[normalem]{ulem}
\usepackage{balance}
\usepackage[bottom]{footmisc}
\usepackage[super]{nth}
\usepackage{subcaption}
\usepackage{cite}
\usepackage{comment}
\usepackage{url}
\usepackage{mwe}

\newfloatcommand{capbtabbox}{table}[][\FBwidth]

\usepackage{pgfplots}
\pgfplotsset{width=7cm,compat=1.8}

\makeatletter
\newcommand{\printfnsymbol}[1]{%
	\textsuperscript{\@fnsymbol{#1}}%
}
\makeatother

\newif\ifdraft
\draftfalse
\drafttrue

\definecolor{orange}{rgb}{1,0.5,0}
\definecolor{violet}{RGB}{70,0,170}

\ifdraft
 \newcommand{\PF}[1]{{\color{red}{\bf PF: #1}}}
 
 \newcommand{\SH}[1]{{\color{blue}{\bf SH: #1}}}
 
 \newcommand{\SR}[1]{{\color{violet}{\bf SR: #1}}}
 
  \newcommand{\LC}[1]{{\color{cyan}{\bf LC: #1}}}

\else
 \newcommand{\PF}[1]{}
 
 \newcommand{\SH}[1]{}
 
 \newcommand{\SR}[1]{}
 
  \newcommand{\LC}[1]{}
 
\fi

\newcommand{\parag}[1]{\vspace{-3mm}\paragraph{#1}}

\def\Vec#1{{\boldsymbol{#1}}}
\def\Mat#1{{\boldsymbol{#1}}}

\newcommand{\mL}{\mathcal{L}}
\newcommand{\mU}{\mathcal{U}}

\newcommand{\bp}{\boldsymbol{p}}

\newcommand{\bI}{\boldsymbol{I}}
\newcommand{\bX}{\boldsymbol{X}}

\def\ie{\emph{i.e.}}

\def\etal{\emph{et al}.}

\definecolor{cadmiumgreen}{rgb}{0.0, 0.42, 0.24}
\usepackage[accsupp]{axessibility}

\definecolor{cvprblue}{rgb}{0.21,0.49,0.74}
\usepackage[pagebackref,breaklinks,colorlinks,citecolor=cvprblue]{hyperref}

\def\paperID{287} 
\def\confName{3DV\xspace}
\def\confYear{2024\xspace}

    \title{Occlusion Resilient 3D Human Pose Estimation}

    \author{Soumava Kumar Roy$^{1}$
    ~~~~~~~Ilia Badanin$^{2}$~~~~~~~Sina Honari$^{3}\thanks{work done while at Computer Vision Lab, EPFL}$~~~~~~~Pascal Fua$^{1}$
    \\
    {$^{1}$ Computer Vision Lab, EPFL, Switzerland} \\
    {$^{2}$ Machine Learning and Optimization Lab, EPFL, Switzerland} \\
    {$^{3}$ Samsung AI Center Toronto} \\
    {\tt\small soumava.roy@epfl.ch ~~ ilia.badanin@epfl.ch ~~ sina.honari@gmail.com  ~~  pascal.fua@epfl.ch}
    }

\begin{document}
\maketitle

\begin{abstract}

Occlusions remain one of the key challenges in 3D body pose estimation from single-camera video sequences. Temporal consistency has been extensively used to mitigate their impact but the existing algorithms in the literature do not explicitly model them. 

Here, we apply this by representing the deforming body as a spatio-temporal graph. We then introduce a refinement network that performs graph convolutions over this graph to output 3D poses. To ensure robustness to occlusions, we train this network with a set of binary masks that we use to disable some of the edges as in drop-out techniques. 

In effect, we simulate the fact that some joints can be hidden for periods of time and train the network to be immune to that. We demonstrate the effectiveness of this approach compared to state-of-the-art techniques that infer poses from single-camera sequences.

\end{abstract}



\section{Introduction}

While there are many compelling algorithms~\cite{Sun18d,Zhou17f,Zhou16b,Li19c,Pavllo19,Chen17h,Tung17a,Kanazawa19b,Chen19g} to capture the pose of a single clearly visible person, modeling partially occluded people in crowded scenes remains an open problem. To address it, several approaches often make strong assumptions~\cite{radwan13a,Rogez17,Iskakov19,Ma21c,roy22a}, which limit their generalization abilities. In this context, exploiting temporal consistency from videos is particularly appealing because it only requires a much weaker assumption of motion continuity. Hence, it has been extensively used to mitigate the impact of occlusions~\cite{Ci19,Cai19,Zeng20,Liu20f,Zou21, hossain18a, Katircioglu18, lee18c} by imposing regularization constraints, but without explicitly accounting for the occlusions. This compromises their robustness to them.


\begin{figure}
    \centering
    \includegraphics[width=1\linewidth,keepaspectratio]{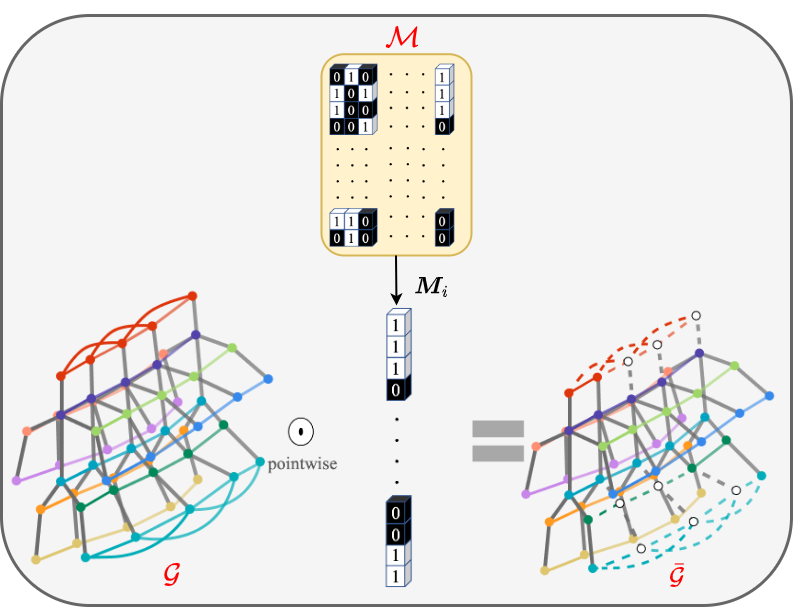}
    \vspace{1mm}
    \caption{ {\bf Graph masking}. $\mathcal{G}$ is our spatio-temporal graph. The solid colors denote graph nodes corresponding to the same joint over time. The gray edges are spatial edges connecting joints seen at the same time while the colored edges are temporal and connect nodes corresponding to the same joint over time. For clarity, we only show spatial and temporal connections for 11 joints. Along the temporal domain, each joint is connected to its temporal neighbor (\ie~$\Delta = 1$). Moreover, we also show temporal connection with $\Delta = 2$ for two joints (\textcolor{red}{\textbf{head}} and \textcolor{cyan}{\textbf{left foot}}) as an example. $\mathcal{M}$ represent the set of binary masks that are used to deactivate some of these edges to create the masked graph $\bar{\mathcal{G}}$, which is then fed to the refinement network. Refer to Section~\textsection~\ref{sec:ref_net} for more details.}
    \label{fig:spatial-temporal-graph-masks}
    \vspace{-0.5cm}
\end{figure}


In this paper, we aim to address this limitation. Given a sequence of consecutive frames, we represent the deforming body as a spatio-temporal graph such as the one depicted by Fig.~\ref{fig:spatial-temporal-graph-masks}, in which spatial edges connect body nodes in one frame to form a skeleton and temporal edges connect the same node at different times. We then introduce a refinement network that performs graph convolutions over this graph to output 3D poses. To ensure robustness to occlusions, we train this network with a set of binary masks that we use to disable some of the edges as in drop-out techniques~\cite{Gal16a,Durasov21a}. In effect, we simulate the fact that some joints can be hidden for periods of time and train the network to be immune to that.

Our contribution is therefore an approach to explicitly modeling occlusions to increase the robustness of a lifting network that produces reliable 3D poses. We demonstrate the effectiveness of our approach on the benchmark Human3.6M~\cite{Ionescu14a}, MPI-INF-3DHP~\cite{Mehta17a} in addition to SportCenter~\cite{roy22a} datasets against several baselines and state-of-the-art methods that use a sequence acquired by a single camera at inference time and can be trained using either single- or multi-view image data.

\section{Related Work}

With recent advances in the field of deep learning, regressing 2D and 3D poses directly from images has become an effective approach to human pose estimation. Some approaches are fully supervised~\cite{Li14e,Li15a,Pavlakos16,Popa17,Sun18d,Tekin16b,Zhou17f,Zhou16b,Rhodin18a,Kocabas19,Mitra20}. Others rely on self-supervision to reduce the required amount of training data~\cite{Li19c,Pavllo19,Tung17a,Chen17h,Kanazawa19b,Chen19g} and make it possible to capture people {\it in the wild}~\cite{Iqbal20}. However, occlusions remain one of the primary challenges, especially in crowded scenes where people may hide each other. Our proposed solution relies on enforcing temporal consistency while explicitly modeling occlusions. As discussed in the remainder of this section, these two aspects have been handled separately in the past but loosely together. In this paper, we take our inspiration from the MC-Droput like method of~\cite{Durasov21a} to enforce temporal consistency while also handling occlusions in a time consistent manner.

\parag{Occlusions.} 

One way to mitigate their influence is to learn camera specific weight for every joint in a calibrated multi camera setup~\cite{Iskakov19,Ma21c,roy22a}. This can be effective but does not generalize well to single, potentially moving, cameras. A more generic approach is to exploit temporal consistency in videos~\cite{lee18c,hossain18a,Pavllo19,cheng19d}. The algorithm of ~\cite{lee18c} learns inter-joint dependencies using an LSTM, whereas~\cite{hossain18a} uses an Recurrent Neural Network (RNN) to enforce temporal smoothness for the detected joints. Unfortunately, these methods do not model the fact that occlusions  exhibit temporally consistent patterns. The approach of~\cite{cheng19d} seeks to enhance the generalization ability of a prediction network by generating augmented pairs of 3D joints and 2D key-points with \emph{arbitrary} per-joint occlusion labels.  However, this arbitrary generation of occlusion labels fails to capture the inherent temporal pattern of the occluded key-points over any given time interval. Our method is in the same spirit but seeks to capture this pattern by using structured occlusion masks of~\cite{Durasov21a}. The works of~\cite{radwan13a,Rogez17} also propose to tackle occlusion by imposing pose priors and kinematic pose constraints on the predicted poses. However, the proposed algorithms fail to impose such constraints in long sequences of images involving many different forms of complex occlusion patterns. As shown in the experiment section, this adversely impacts performance even when using training sets that contain substantial amounts of occlusions. 

\parag{Spatio-Temporal Consistency.} 

Recently, Graph  Convolution Neural Networks (GCNNs)~\cite{Defferrard16,Kipf16,Levie2018}  have become increasingly popular to exploit the skeletal structure of human joints~\cite{Ci19,Cai19,Zeng20,Liu20f,Zou21,Xu21b}. In~\cite{Cai19}, predicted 2D joints locations are assembled into a graph and a network learns to combine different local features corresponding to different body parts to produce a global embedding, which is eventually used to decode the final 3D pose in a target frame. Similarly, the algorithm of~\cite{Zeng20} splits and recombines part-wise feature representations to learn inter-joint dependencies across different body parts. In~\cite{Xu21b} weight and affinity modulation methods are introduced to learn a disentangled representation of a spatial skeletal graph. In~\cite{Cai19,Liu20f,Zou21}  structured weight matrices are used to model inter-joint relationships, thus reducing the overall computational complexity of the framework. However, none of these methods explicitly handle occlusions.

\begin{figure*}
    \includegraphics[trim=5cm 0 4cm 0, clip,width=1\textwidth]{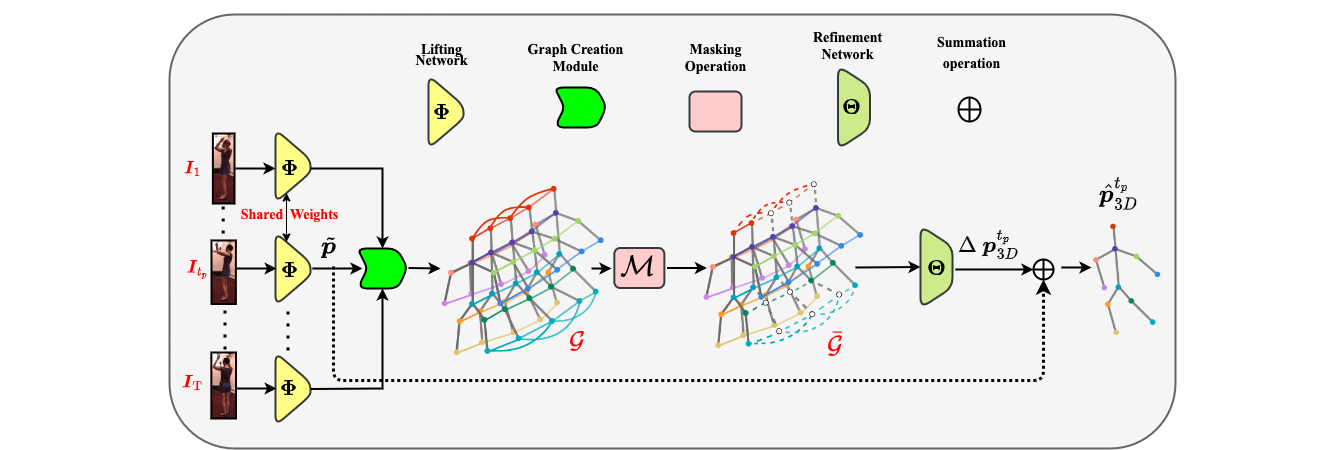}
    \caption{{\bf Our approach.} 3D joint coordinates are extracted from individual images by the lifting net $LNet_{\Mat{\Phi}}$ and become the nodes of a spatio-temporal graph $\mathcal{G}$. Some of its edges are masked to produce a reduced graph $\bar{\mathcal{G}}$. It is fed to a refinement network $RNet_{\Mat{\Theta}}$ that returns the pose in the selected target frame $t_p$. The masking operation is depicted by Fig.~\ref{fig:spatial-temporal-graph-masks} (Refer to Section~\textsection~\ref{sec:lift_net} and~\ref{sec:ref_net} for more details).}
    \label{fig:overview}
\end{figure*}


\section{Method}

We train an occlusion-robust network to estimate 3D human poses from image sequences acquired using a single camera. As shown in Fig.~\ref{fig:overview}, we first detect 2D joints in individual images. They are mapped to 3D by a {\it lifting network} and become the nodes of a spatio-temporal graph that represents the body deforming over time. This graph is fed to a {\it refinement network} that produces a 3D pose for a target frame in the middle of the sequence. A key component of this approach is the set of structured binary masks depicted by Fig.~\ref{fig:spatial-temporal-graph-masks}. They are used to deactivate some edges in the graph before it is fed to the refinement network during training. This increases robustness to occlusions. 


\subsection{Notations} 
\label{sec:notations}

Let $\mL = \{\bI^{l}, \bp^l_{3D}\}_{l=1}^{\textrm{N}_\textrm{L}}$ be a monocular sequence of labeled $\textrm{N}_\textrm{L}$ RGB images $\Mat{I}^{l} \in \mathbb{R}^{h \times w \times 3}$, featuring a person.  $\bp^l_{3D}$ denotes the ground-truth 3D pose of the person's body and $l$ is the image index. $\bp^l_{3D}$ stands for a full set of $\textrm{N}_\textrm{J}$ body joints $\{\bX^l_j\}_{j=1}^{\textrm{N}_\textrm{J}}$, where $\bX^l_j \in \mathbb{R}^3$ denotes the 3D coordinates of joint $j$ in image  $l$. 

We use $\mL$ for supervised single-view learning. For unsupervised learning, let $\mU = \{\bI^{u,v}\}_{u=1,v=1}^{\textrm{N}_\textrm{U},\textrm{N}_\textrm{V}}$ be a longer multi-view sequence of $\textrm{N}_\textrm{U}$ images captured in $\textrm{N}_\textrm{V}$ views simultaneously but without associated body poses. This represents an important training scenario because setting up multiple cameras is a relative inexpensive way to provide supervision without having to annotate, which is far more laborious. In the remainder of this paper, we drop the $l$ and $u$ indices when there is no ambiguity.

A human body deforming in a sub-sequence of length $\textrm{T}$ can be represented by a graph $\mathcal{G}(\mathcal{V}, \mathcal{E})$ whose vertices $\mathcal{V} = \left\{v_{tj}~~|~~t = 1, 2, \cdots \textrm{T};~~j = 1, 2, \cdots \textrm{N}_\textrm{J}  \right\}$ are the body joints $\bX^l_j$ corresponding to T frames and whose edges $\mathcal{E}$ are both spatial and temporal. In Fig.~\ref{fig:spatial-temporal-graph-masks}, the spatial edges are shown in gray; they define the skeleton within each temporal frame. The temporal edges are shown in color and connect joints over time. In practice, we connect a joint in image $l$ to all its temporal neighbors from frame $l-\frac{T}{2}$ to $l+\frac{T}{2}$ with a temporal stride $\Delta = \left\{1, 2, 3 \cdots \Delta_{max} \right\}$, where $\Delta_{max}$ is the maximum temporal stride we consider in the window of size T. 
We denote by $\Mat{A} = [a_{ij}]$ the {\it adjacency matrix} of the graph such that $a_{ij}=1$ if nodes $i$ and $j$ are connected and 0 otherwise.~\textrm{N} = $\textrm{T}\!\ast \!\textrm{N}_\textrm{J}$ represent the total number of joints (or noded) in the graph $\mathcal{G}$. 
\subsection{Lifting Network ($LNet_{\Mat{\Phi}}$)}
\label{sec:lift_net}

We start from 2D joint detections obtained by an off-the-shelf joint detector~\cite{roy22a}. In each image $\Mat{I}$, the lifting network $LNet_{\Mat{\Phi}}$, with trainable weights $\Phi$ maps these 2D detections to 3D points that we treat as estimates $\tilde\bp_{3D}$ of the true body pose  $\bp_{3D}$. This is depicted on the left side of Fig.~\ref{fig:overview}. Under the assumption that intrinsic camera parameters $\Mat{K}$ are known, we use the standard approach of~\cite{Sun18d, Pavlakos18a, Iqbal18,Iqbal20} to map the predictions of  $LNet_{\Phi}$ into coordinates expressed into the camera coordinate system. These are expressed in terms of $d^{root}$, the distance of the root joint to the camera and $d^{rel}_j$ the distance difference for each joint with respect to that of the root. We write
%

\begin{equation}
\label{eqn:2d_3d_lift_depth}
\begin{aligned}
    (d^{root}+d^{rel}_j)\begin{pmatrix}
           x_j \\
           y_j \\
			1
         \end{pmatrix} 
         &
         = \Mat{K} 
         \begin{pmatrix}
           X^C_j \\
           Y^C_j \\
			(d^{root}+d^{rel}_j)
         \end{pmatrix}
        \\ &
         = \Mat{K} \left[ \Mat{R}
         \begin{pmatrix}
           X_j \\
           Y_j \\
		   Z_j
         \end{pmatrix}    
         + \Vec{t}      
          \right]~~~~ , 
\end{aligned}
\end{equation}

where $X_j^C,~Y_j^C$ denote the first two 3D coordinates of joint $j$ in the camera coordinate system and $x_j,~y_j$ denote their projection on to the un-distorted image space respectively. $(X_j,Y_j,Z_j)$ is the joint in the world coordinate and $\Mat{K} \in \mathbb{R}^{3 \times 3}$, $\Mat{R} \in \mathbb{R}^{3 \times 3}$, and $\Vec{t} \in \mathbb{R}^{3 \times 1}$ are the intrinsic matrix, rotation matrix, and translation vector of the camera respectively. As in previous work~\cite{Rhodin18a, Martinez17a, Sun18d}, we use the ground-truth value of $d^{root}$. However an analytical approximation can be obtained if necessary~\cite{Iqbal18}. 

The lifting network $LNet_{\Mat{\Phi}}$ is trained to predict these distances by learning a set of weights $\Phi$. Within each continuous sequence of frames the resulting 3D points are then assembled into a spatio-temporal graph  $\mathcal{G}$.

\subsection{Refinement Network  ($RNet_{\Mat{\Theta}}$)}
\label{sec:ref_net}

The spatio-temporal graph $\mathcal{G}$ built on the predictions of $LNet$ is a raw assembly of lifted 2D poses and one can expect many of its 3D nodes to be inaccurate, especially when there are occlusions. The problem is compounded by the fact that $LNet$  does not deliver an uncertainty estimate for its predictions.  We therefore feed it to  a refinement network $RNet_{\Mat{\Theta}}$ whose role is to exploit temporal consistency constraints to output an accurate 3D pose for the central frame of the sequence it has been constructed from,  as shown on the right side of Fig.~\ref{fig:overview}.

We will refer to this frame as the target frame. Our challenge is to make $RNet$ robust to occlusions. 

\subsubsection{Robustness to Occlusions}

In practice, occlusions hide parts of the body. Hence, some of the 2D detections and the resulting 3D lifted points that $LNet$ produces are bound to be incorrect. To make our $RNet$ robust to this, we take our inspiration from MC-Dropout~\cite{Gal16a}, which must be able to ignore some nodes in $\mathcal{G}$ and yet deliver the correct target pose. 

The original MC-Dropout idea involves dropping weights at random by introducing random binary masks that define which weights are used or ignored. Unfortunately, this does not reflect the true behavior of occlusions. Occluded body joints tend to be spatial neighbors and the occlusions exhibit temporal consistency. Hence the masks should be correlated both spatially and temporally. Therefore, as in~\cite{Durasov21a}, we define a fixed set of pre-computed parameterized binary masks $\mathcal{M} = \left [ \Mat{M}_i \right ]_{i=1}^{\textrm{N}_{\textrm{M}}} = \Omega \left(\textrm{N}_{\textrm{M}}, \beta, \alpha \right)$, where $\Omega$ is the parameterized mask generating function, $\textrm{N}_{\textrm{M}}$ is the total number of masks, $\Mat{M}_i=\left\{0, 1\right\}_{j=1}^{\textrm{N}}$, $\beta$ denotes the number of 1's in each mask, and $\alpha$ controls the amount of overlap between masks. The larger the overlap, the higher the chances that the same nodes will be dropped in subsequent frames, thus enforcing a degree of temporal consistency that depends on $\alpha$. 

As shown in Fig.~\ref{fig:spatial-temporal-graph-masks}, during training, we randomly pick masks from the generated set in $\mathcal{M}$ and use them to disconnect some edges in $\mathcal{G}$. This produces subgraphs $\bar{\mathcal{G}}$, which we feed to $RNet$.

\subsubsection{Architecture and Losses}

\begin{figure}[!t]
    \centering
    \includegraphics[trim=0 0 5cm 0, clip,width=1\linewidth,keepaspectratio]{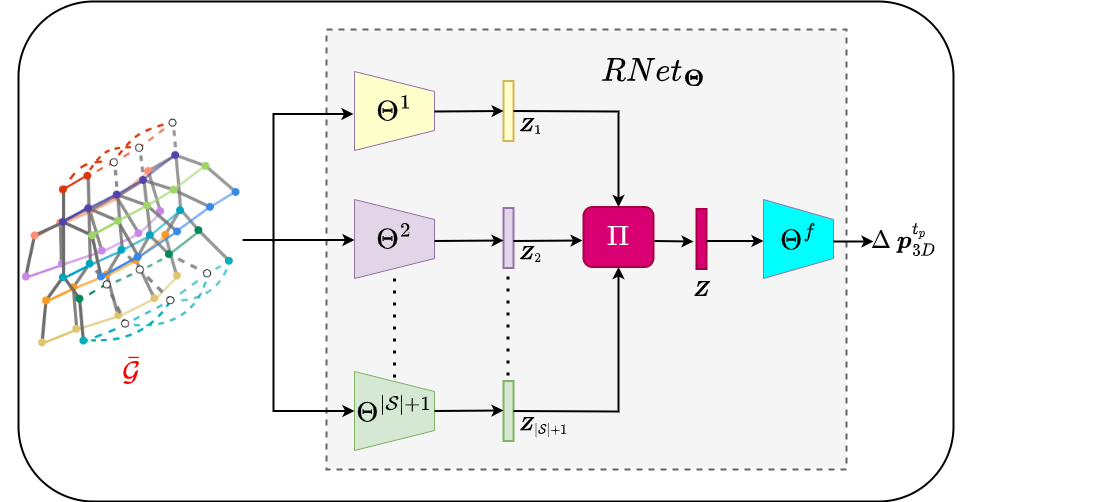}
    \vspace{-4mm}
    \caption{{\bf Architecture of the refinement network.} $RNet_{\Mat{\Theta}}$ is a set of GCNNs that operates on the masked graph $\bar{\mathcal{G}}$. Each \emph{relationship-specific} GCN is trained on a different set of connections between the joints, which are eventually fused and processed by the \emph{embedding-fusion} network with parameters $\Mat{\Theta}^f$. 
    }
    \label{fig:gcn_operation}
\end{figure}

We implement $RNet_{\Theta}$ as a Graph Convolutional Neural Network (GCNN)~\cite{Defferrard16,Kipf16,Levie2018}. In its successive layers, a feature vector is assigned to each node of the graph and averaged with its neighbors according to a set of learned weights in $\Theta$. The results are then pooled as in ordinary CNNs. Finally, the output is fed to a linear layer with output dimensionality of $\textrm{N}_{\textrm{J}}$ that predicts the incremental root-relative depth $\Delta d^{rel}_j$ for each joint of the target frame $t_p$. These are added to the predictions of $LNet$ to yield the final refined depth of all joints, which are transformed to $\hat{\Mat{p}}_{3D}^{t_p}$ using Eq.~(\ref{eqn:2d_3d_lift_depth}) for the target frame $t_p$, as shown in Fig.~\ref{fig:gcn_operation}. 

The spatio-temporal graph~$\mathcal{G}$ should encode the spatial and temporal connections between the joints in the temporal window of size T. As the 3D joint positions are correlated, we therefore need to learn these correlations within T. Thus, we connect the joints across time with different values of the temporal stride $\Delta$, which are intended to model relationships at different temporal scales. Let $\mathcal{S}$ be the set of all temporal connections generated using different values of $\Delta$s. Hence, we have $\left|\mathcal{S}\right|$ + 1 different edge types, $|\mathcal{S}|$ temporal types and \emph{one} spatial type. Each type is processed by a separate \emph{relationship-specific} GCNN that combines information about the joint using both spatial and temporal neighbors.

\parag{Relation-Specific GCN.}

The initial value of node $v_{tj}$ is set to $d^{rel}_j$, predicted by the lifting network as shown in Eq.~(\ref{eqn:2d_3d_lift_depth}) for the frame $t \in \left [ 1, 2, ... \textrm{T} \right ]$. This  yields the initial feature matrix $\Mat{H} \in \mathbb{R}^{\textrm{N}}$. For masking purposes, $\Mat{H}$ is multiplied by one of the binary masks $\Mat{M}_i \in \mathcal{M}$ to obtain the feature matrix $\bar{\Mat{H}} = \Mat{H} \odot \Mat{M}_i$ of the masked graph $\bar{\mathcal{G}}$. This is possible because $\bar{\mathcal{G}}$ has the same structure as $\mathcal{G}$, with some of its node features set to $0$ according to the location of zeros in $\Mat{M}_i$.


As shown in Fig.~\ref{fig:spatial-temporal-graph-masks}, in the graph $\mathcal{G}$ (or $\bar{\mathcal{G}}$) we consider $\left | \mathcal{S} \right |$ + 1 different connections between the joints, which are individually processed by a \textbf{r}elationship-specific GCN $RNet_{\Mat{\Theta}^\textbf{r}}$. Each of these $RNet_{\Mat{\Theta}^\textbf{r}}$ takes as input the initial feature matrix $\bar{\Mat{H}}$ and outputs $\Mat{Z}_\textbf{r} \in \mathbb{R}^{\bar{\textrm{d}}}$ for every node $v_{tj}$. Thereafter, we pass $\left \{ \Mat{Z}_\textbf{r} \right \}_{\textbf{r}=1}^{\left | \mathcal{S} \right |+1}$ through a pooling operation $\Pi$ to obtain the final feature matrix $\Mat{Z} = \Pi ( \left \{ \Mat{Z}_r \right \} ) \in \mathbb{R}^{\textrm{N} \times \bar{\textrm{d}}}$ for the graph $\bar{\mathcal{G}}$. $\Mat{Z}$ is further passed through an additional \emph{embedding-fusion} neural network $RNet_{\Mat{\Theta}^\textbf{f}}$ that returns $\Delta\Mat{p}^{t_p} = \left\{\Delta d_j^{rel} \right\}_{j=1}^{\textrm{N}_\textrm{J}}$, which is further added to $d_j^{rel}$ predicted by the lifting network $LNet_{\Mat{\Phi}}$. The summed value yields the estimated 3D pose $\hat{\Mat{p}}^{t_p}_{3D}$ using Eq.~(\ref{eqn:2d_3d_lift_depth}) for the frame $t_p$. Fig.~\ref{fig:gcn_operation} shows a diagrammatic representation of operations by the refinement network $RNet_{\Mat{\Theta}}$.






\parag{Training Losses.} 

For supervised learning exploiting the labeled image set ${\cal L}$, we take the loss for a sequence of length $\textrm{T}$ to be
\begin{equation}
\label{eqn:loss_refine_net}
    L(\Mat{\Phi}, \Mat{\Theta}) = L_{\textrm{MSE}} \Big(\hat{\Mat{p}}^{t_p}_{3D}, \Mat{p}^{t_p}_{3D} \Big)~~,
\end{equation}
where $L_{\textrm{MSE}}$ is mean squared error loss, while $\hat{\Mat{p}}_{3D}^{t_p}$ and $\Mat{p}_{3D}^{t_p}$ are the estimated and ground-truth poses in the target frame $t_p$. 
To exploit the unlabeled image multi-view set ${\cal U}$, we replace the missing ground-truth poses $\Mat{p}_{3D}^{t_p}$ in Eq.~(\ref{eqn:loss_refine_net}) by 3D pseudo labels, which are generated by running the robust weighted triangulation scheme of~\cite{roy22a} on the joints detected by the 2D pose estimator.

\subsection{Implementation Details}
\label{sec:implementation_details}

\noindent\textbf{Lifting Network.} $LNet_{\Mat{\Phi}}$ consists of a multi-layer perceptron network with 2 hidden layers with a hidden size of 2048. The first two layers are composed of a linear layer, each followed by ReLU activation and 10\% dropout while the last layer is a linear layer with the output dimension set to $\textrm{N}_\textrm{J}$. Given an input image $\Mat{I}$, $LNet_{\Mat{\Phi}}$ takes as input the 2D predictions of a 2D pose estimator model $\hat{\bp} \in \mathbb{R}^{\textrm{N}_\textrm{J} \times 2}$ as the input. We implement the 2D pose estimator model of~\cite{roy22a} to extract the initial 2D key-points, which are further normalized in the range of $\left [ -1, 1 \right ]$ to yield $\hat{\bp}$.
 
\parag{Refinement Network.} The value of $\textrm{T}$, \ie~ the number of frames in each window, is set $31$. This results in $31 \ast \textrm{N}_\textrm{J}$ nodes in the graph~$\mathcal{G}$, where $\textrm{N}_\textrm{J}$ is the number of estimated joints. We choose $\Delta = \{1, 3, 5, 7\}$ resulting in $4$ types of temporal edges and $1$ of spatial ones in $\mathcal{G}$. In $~\mathcal{M}$, we have set $\textrm{N}_\textrm{M}$ and $\alpha$ to $32$ and $1.8$ respectively. This results in $\beta = 292$ in each $\Mat{M}_i \in \mathcal{M}$ when $\textrm{N}_\textrm{J}$ is set $17$ for Human 3.6M~\cite{Ionescu14a} and MPI-INF-3DHP~\cite{Mehta17a} datasets, and 223 when we use 13 joints for SportCenter dataset~\cite{roy22a} in our experiments.

Each \emph{relationship-specific} network $RNet_{\Mat{\Theta}^\textbf{r}}$ consists of $2$ hidden linear layers with the hidden sizes set to $16$ and $32$ respectively, each followed by ReLU activation. The final layer is a linear layer with its output dimension set to $64$, thus resulting in $\Mat{Z}_{\textbf{r}} \in \mathbb{R}^{\textrm{N} \times 64}$. We have assigned $\Pi$ to \textbf{max} pooling operation to obtain the $\Mat{Z}$. The \emph{embedding-fusion} neural network $RNet_{\Mat{\Theta}^\textbf{f}}$ is composed of a single linear layer with the output dimension set to $\textrm{N}_{\textrm{J}}$. We always predict the 3D pose of the central frame (~\ie~$t_p =15)$ present in the time window $\textrm{T}$.

We use the Adam optimizer~\cite{Kingma15} for optimizing all the parameters of $LNet_{\Mat{\Phi}}$ and $RNet_{\Mat{\Theta}}$. As a warm startup for the first 1K iterations, we train $\Mat{\Phi}$ with a constant learning rate of $10^{\textrm{-4}}$ on the labeled set $\mL$ by minimizing the following the loss function
\begin{equation}
\label{eqn:loss_lift_net}
    L(\Mat{\Phi}) = L_{\textrm{MSE}} \Big(\tilde{\Mat{p}}_{3D}, \Mat{p}_{3D} \Big) \; ,
\end{equation}
where $\tilde{\Mat{p}}_{3D}$ are the 3D predictions by $LNet_{\Mat{\Phi}}$. Thereafter we train both set of parameters $\Mat{\Phi}$ and $\Mat{\Theta}$ by minimizing the loss of Eq.~(\ref{eqn:loss_refine_net}) with a constant learning rate of 5$ \times 10^{\textrm{-5}}$ and 1 $\times 10^{\textrm{-4}}$ respectively.

\section{Experiments}

We perform experiments in the \textbf{semi-supervised} learning setup to validate the performance of our proposed method. More specifically, training of the networks is done on the labeled set $\mathcal{L}$ and the unlabeled multi-view set $\mathcal{U}$\footnote{We also provide results for the \textbf{fully-supervised} setup in the supplementary material.}. In both cases, the inference occurs on single-view images.



\subsection{Datasets and Metrics}
\label{sec:datasets}

We briefly describe the datasets and metrics we use to perform our experiments.

\parag{Human 3.6M~\cite{Ionescu14a}.}

It is the most widely used indoor dataset for single and multi-view 3D pose estimation. It consists of $3.6$ million images captured form $4$ calibrated cameras. As in most published papers, we use subjects S$1$, S$5$, S$6$, S$7$, S$8$ for training and S$9$, S$11$ for testing. In the semi-supervised setup we follow two different training protocols:
\begin{enumerate}

    \item $\mathcal{L}$ comprises every 10$^{\textrm{th}}$ frame sampled uniformly from the training set, while the remaining images are treated as being unlabeled~\cite{Kundu20} and form the set  $\mathcal{U}$.  

    \item As in~\cite{Rhodin18a, Pavlakos19b, Li18c}, the supervised set $\mathcal{L}$ comprises images of subject S$1$, while images of the remaining training subjects constitute the multi-view dataset $\mathcal{U}$.

\end{enumerate}

\parag{MPI-INF-3DHP~\cite{Mehta17a}.}

It includes both  indoor and  outdoor images for single person 3D pose estimation. It features $8$ subjects performing $8$ different actions, which are recorded using $14$ different cameras, thereby covering a wide range of  3D poses and viewing angles. It contains both indoor and complex outdoor scenes, covering a wide range of actions ranging from walking, sitting to challenging exercise poses with dynamic motions which undergo a considerable amount of occlusion. We follow the standard protocol and use the 5 chest-height cameras only. In the semi-supervised setup,  we use the images of subject S$1$ to form  $\mathcal{L}$ and the others are taken to belong to $\mathcal{U}$.

\begin{figure}
    \centering
    
    \begin{subfigure}[b]{0.3\linewidth}        
        \centering
        \includegraphics[width=\linewidth]{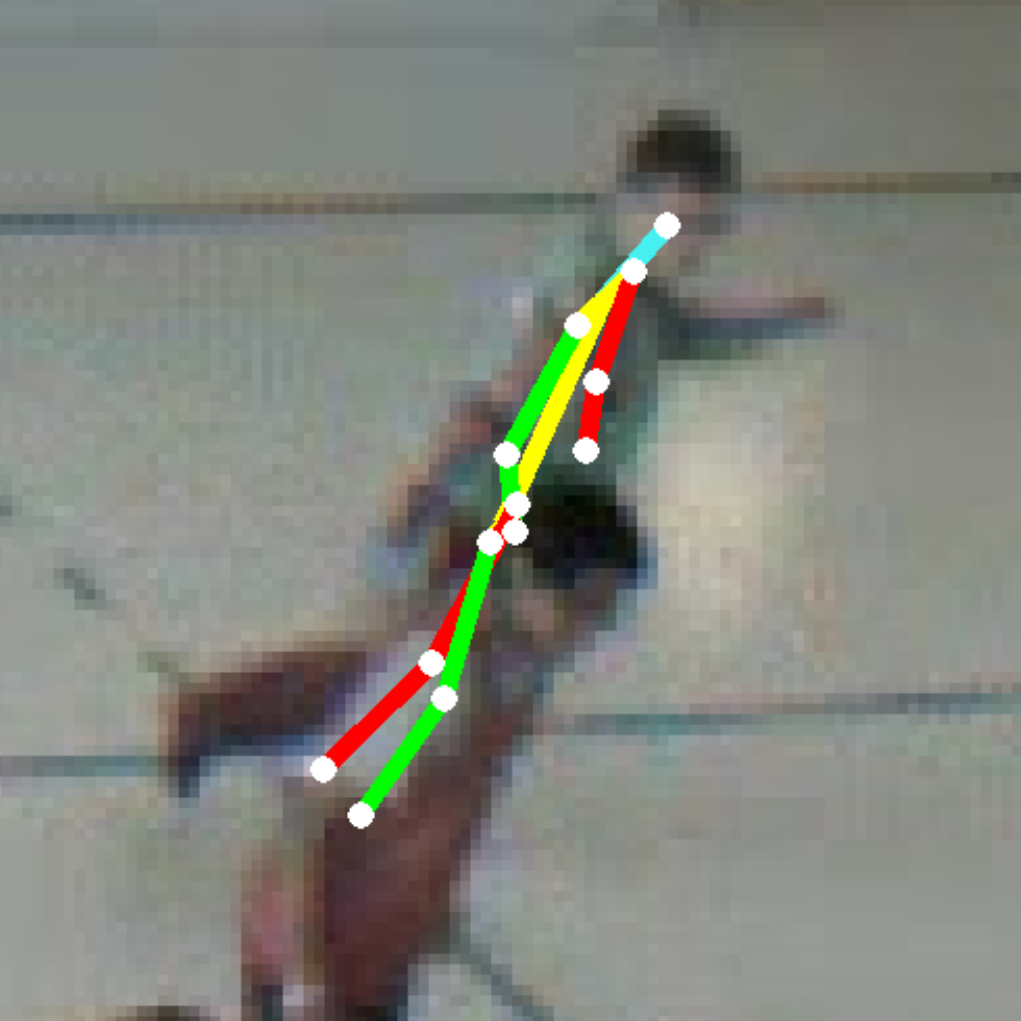}
    \end{subfigure}
    \begin{subfigure}[b]{0.3\linewidth}        
        \centering
        \includegraphics[width=\linewidth]{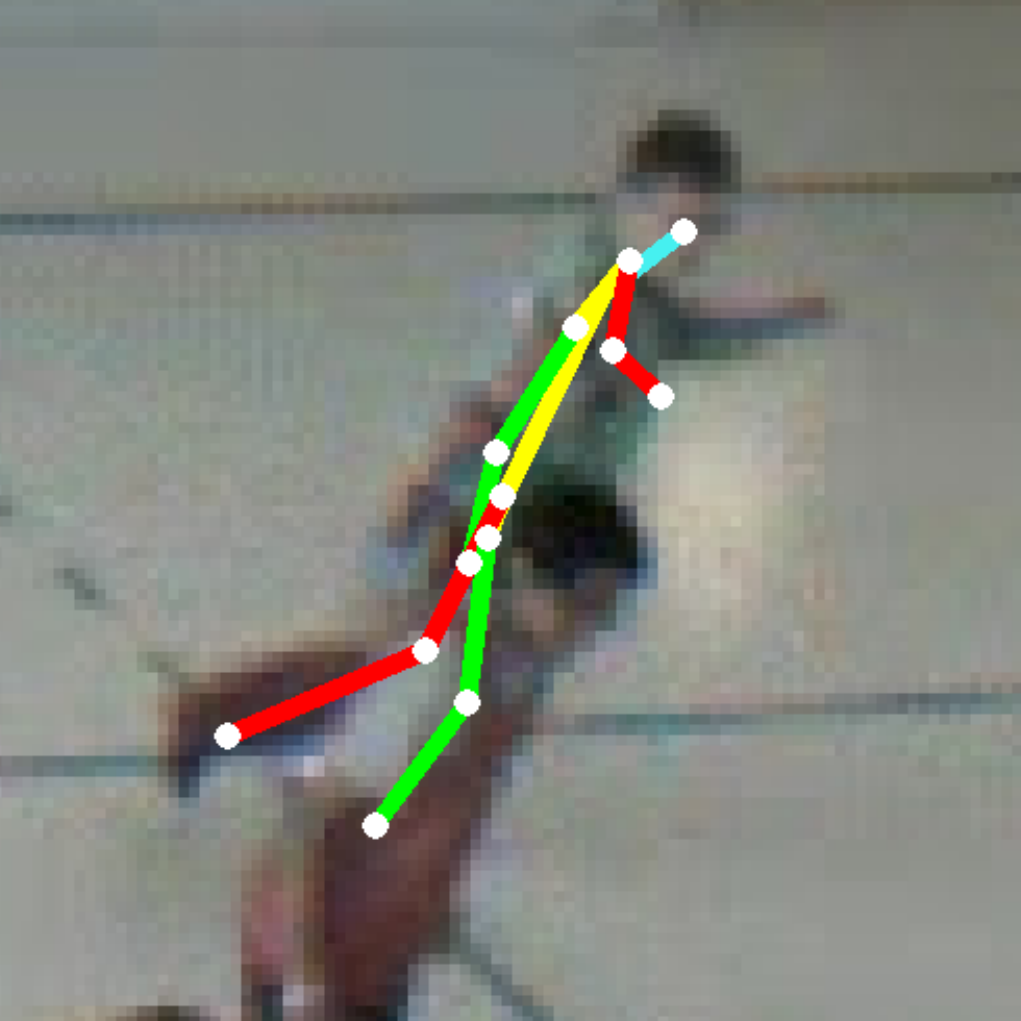}
    \end{subfigure}
    \begin{subfigure}[b]{0.3\linewidth}        
        \centering
        \includegraphics[width=\linewidth]{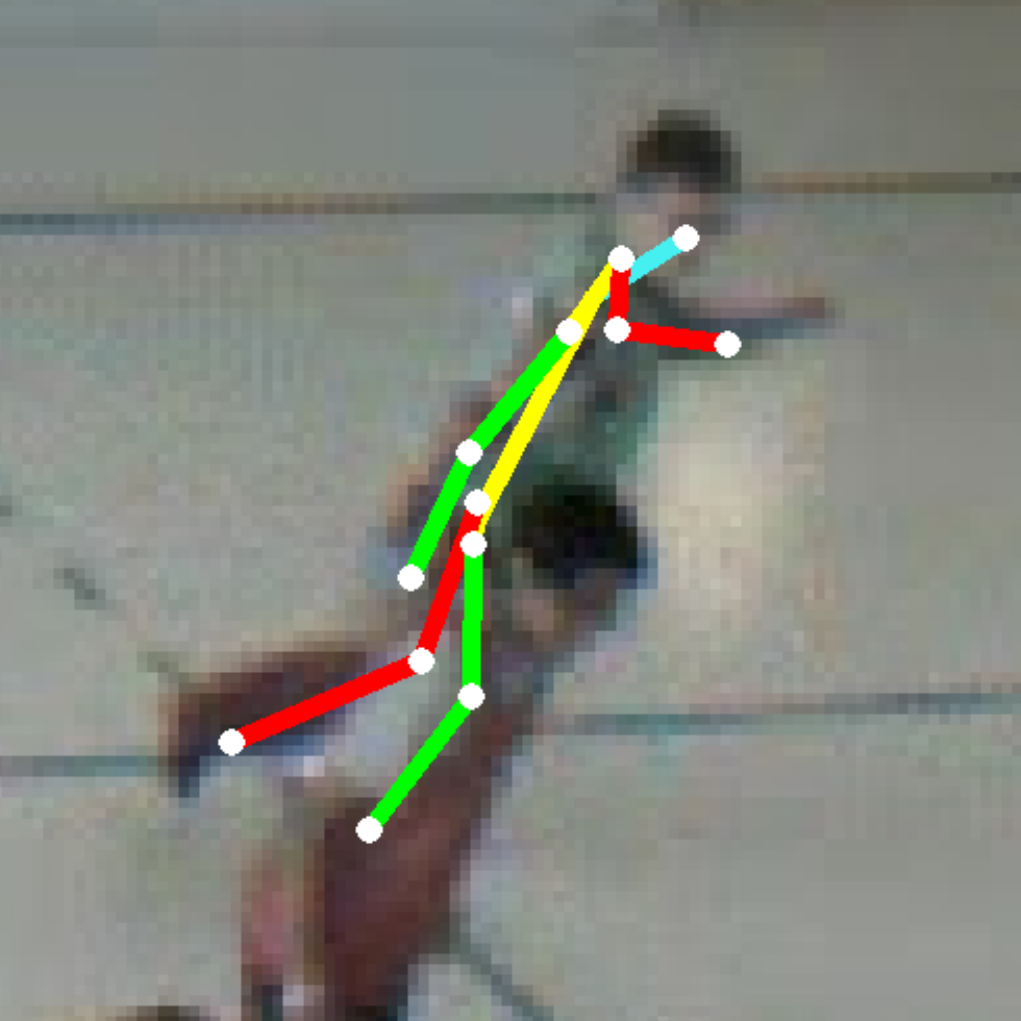}
    \end{subfigure}
    \\

    \begin{subfigure}[b]{0.3\linewidth}        
        \centering
        \includegraphics[width=\linewidth]{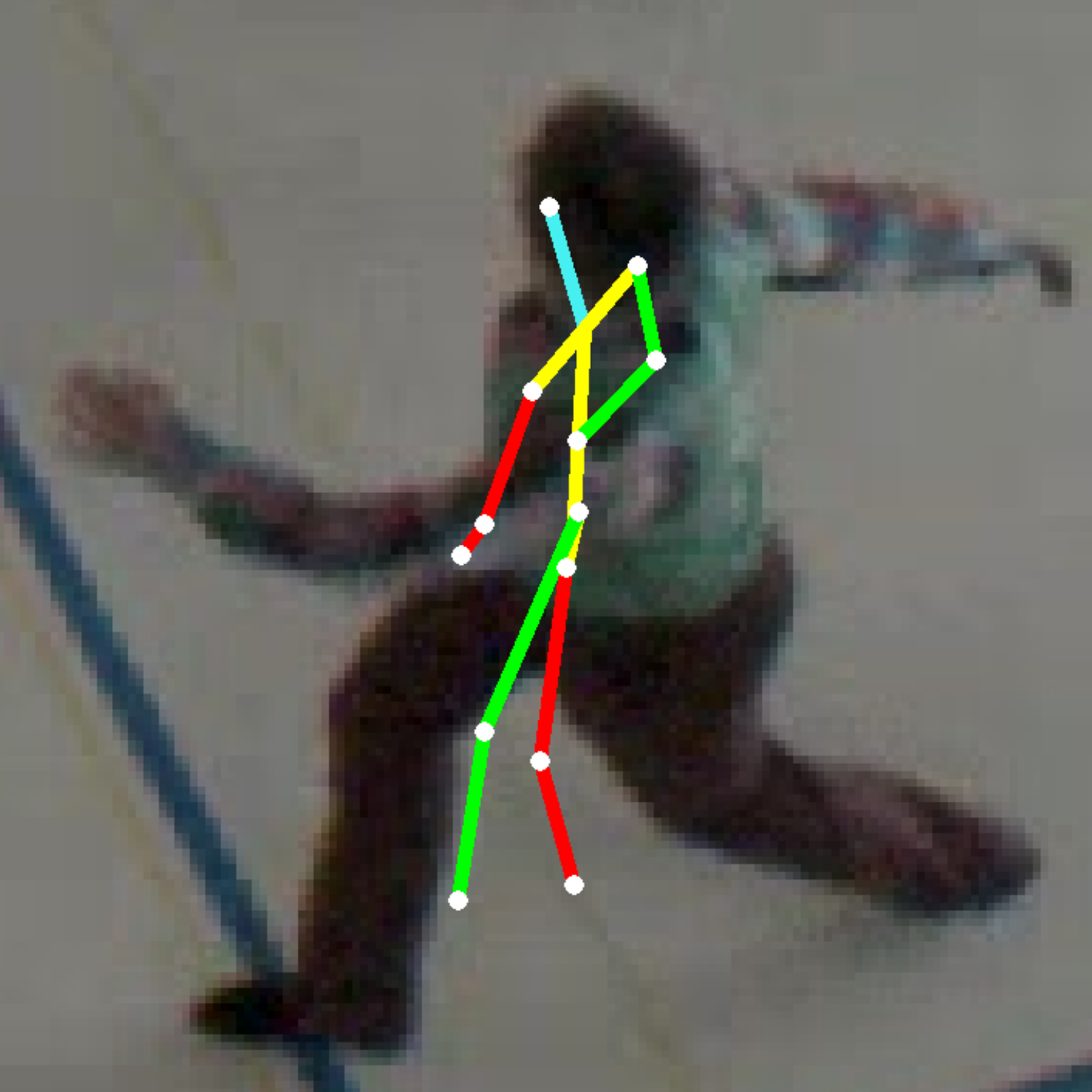}
    \end{subfigure}
    \begin{subfigure}[b]{0.3\linewidth}        
        \centering
        \includegraphics[width=\linewidth]{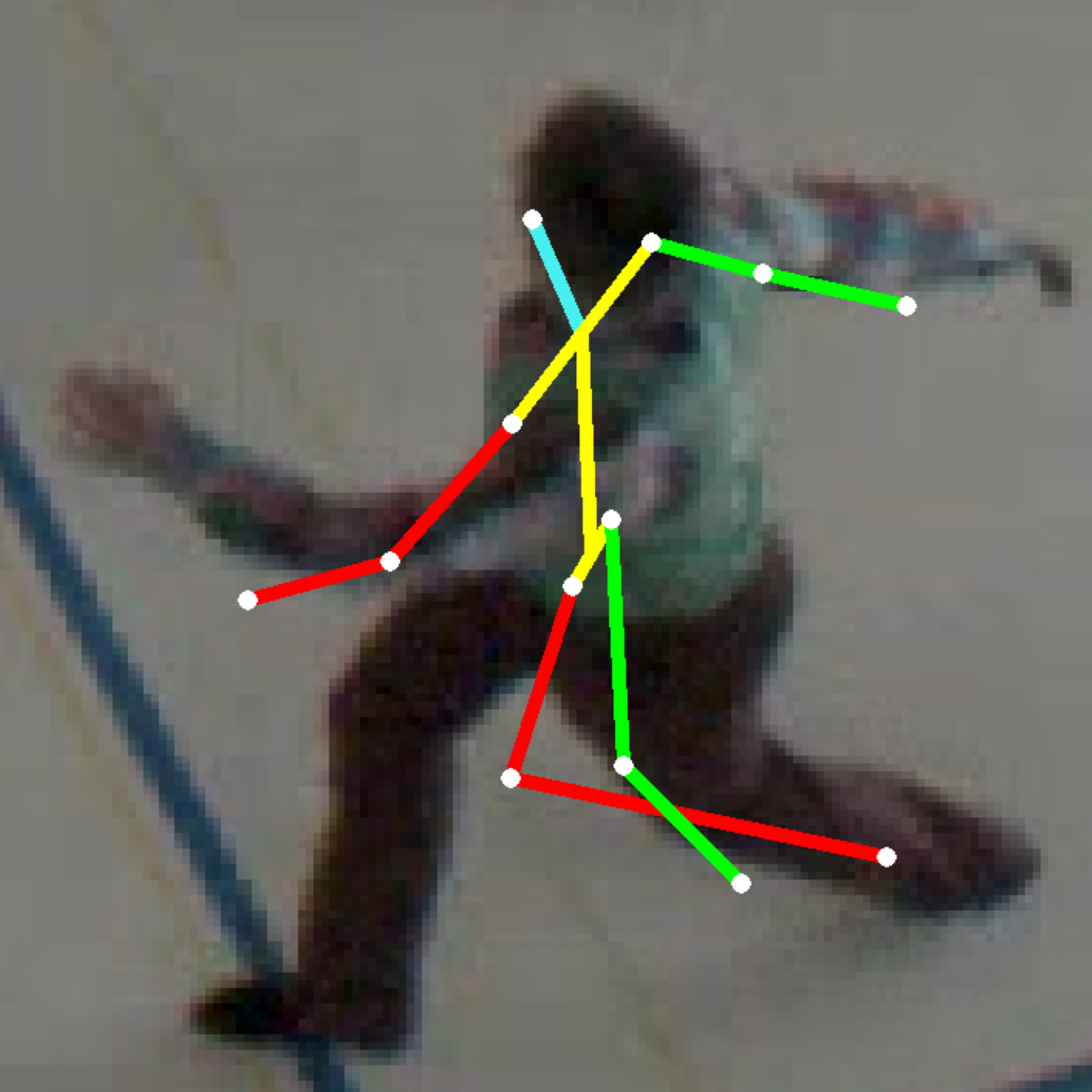}
    \end{subfigure}
    \begin{subfigure}[b]{0.3\linewidth}        
        \centering
        \includegraphics[width=\linewidth]{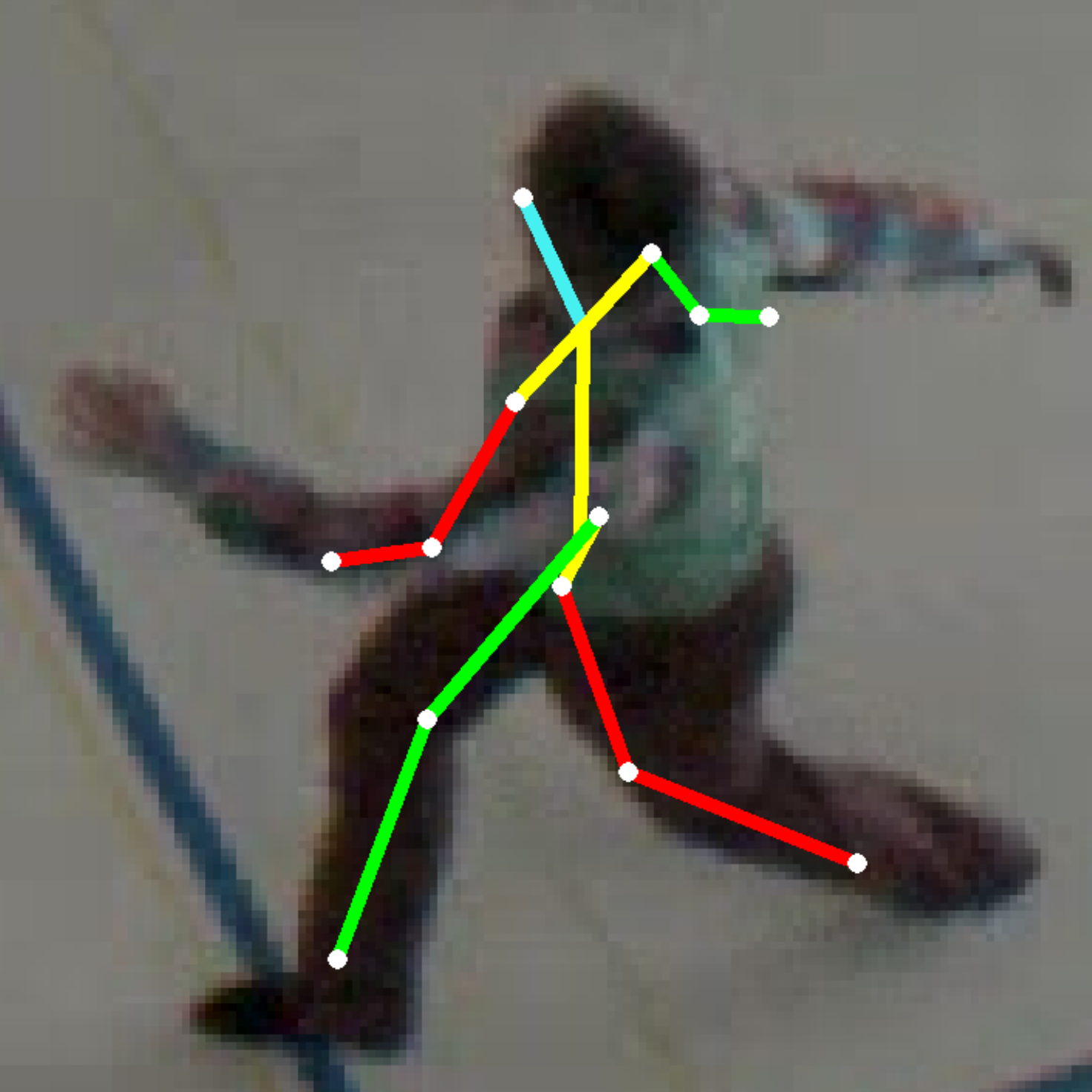}
    \end{subfigure}
    \\

\begin{subfigure}[b]{0.3\linewidth}        
        \centering
        \includegraphics[width=\linewidth]{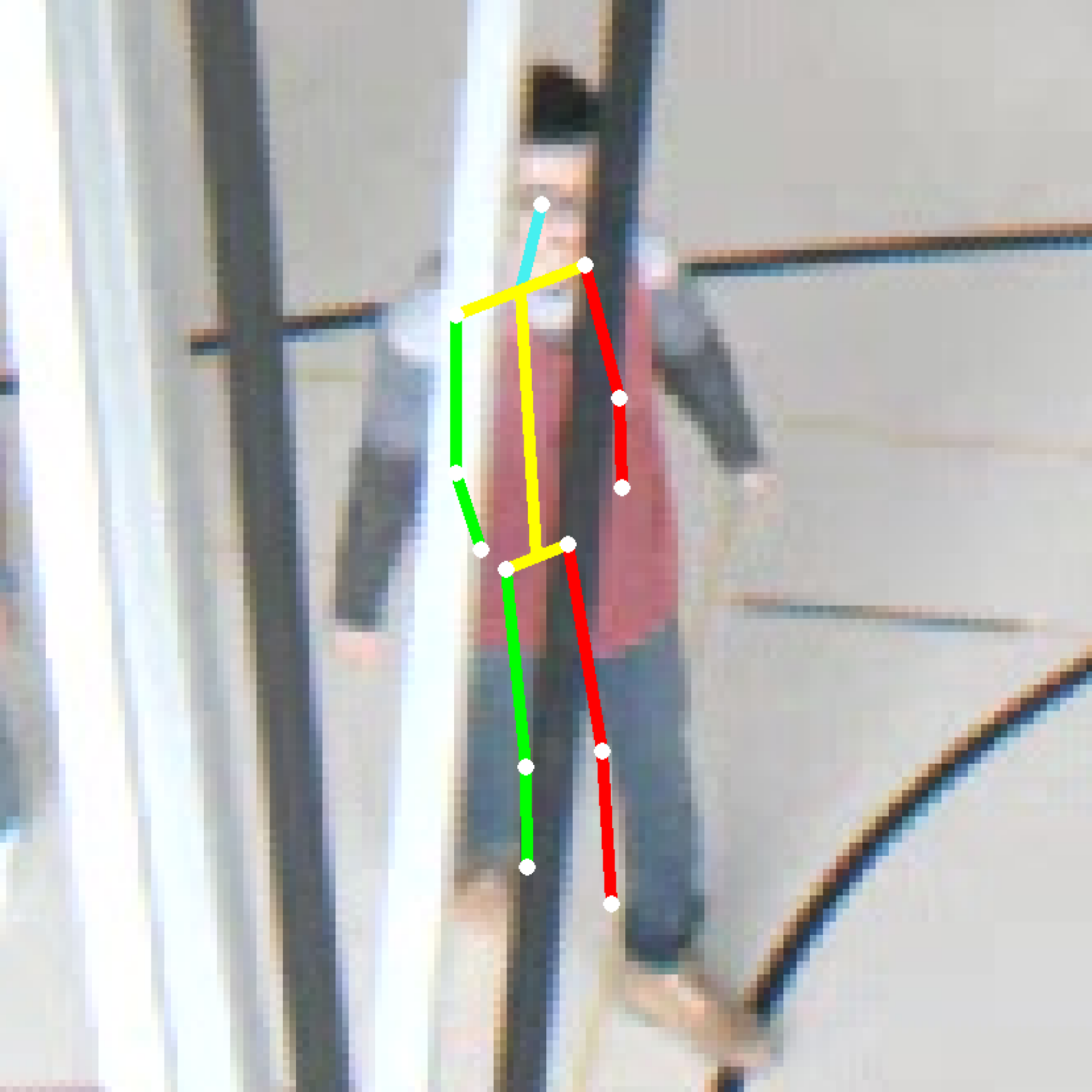}
    \end{subfigure}
    \begin{subfigure}[b]{0.3\linewidth}        
        \centering
        \includegraphics[width=\linewidth]{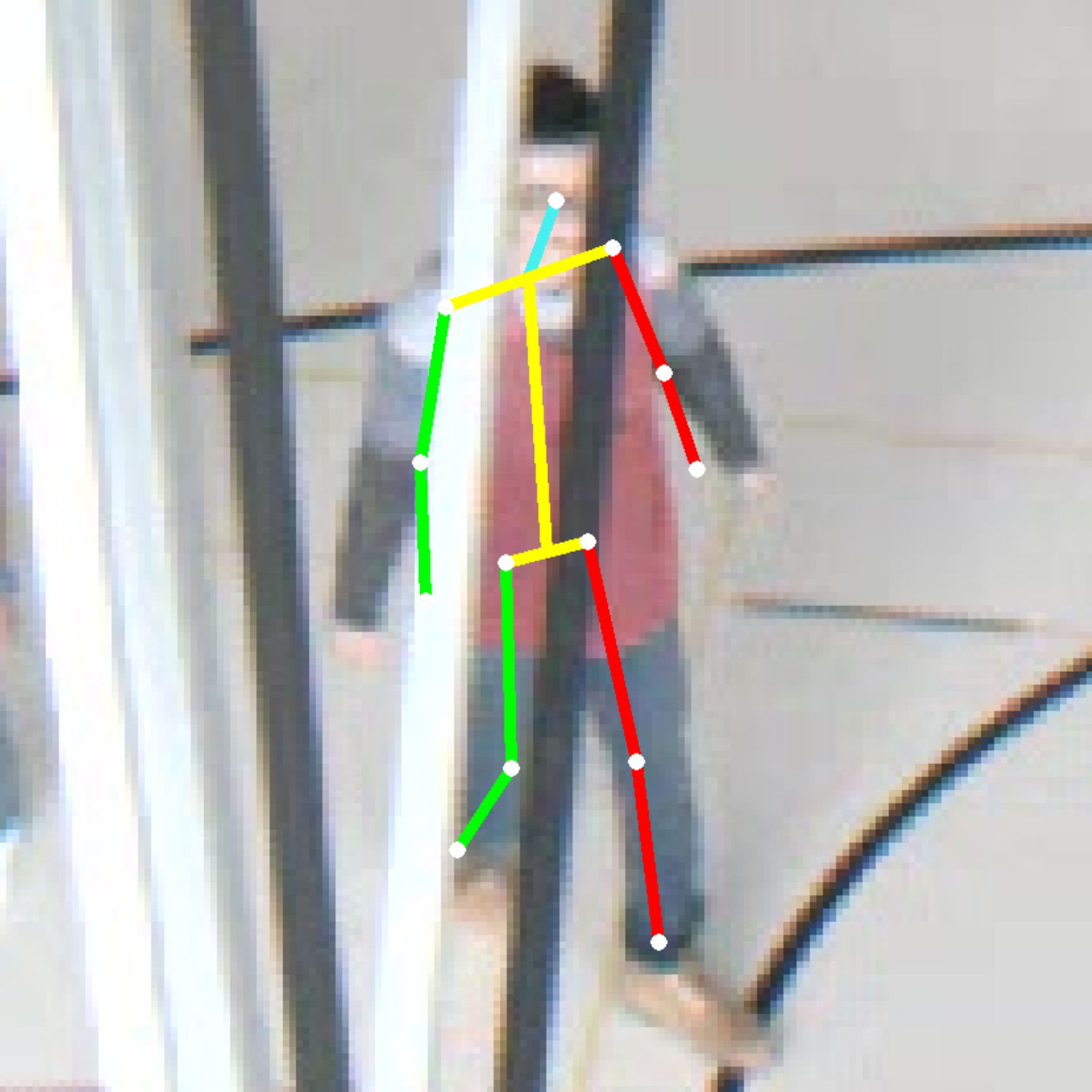}
    \end{subfigure}
    \begin{subfigure}[b]{0.3\linewidth}        
        \centering
        \includegraphics[width=\linewidth]{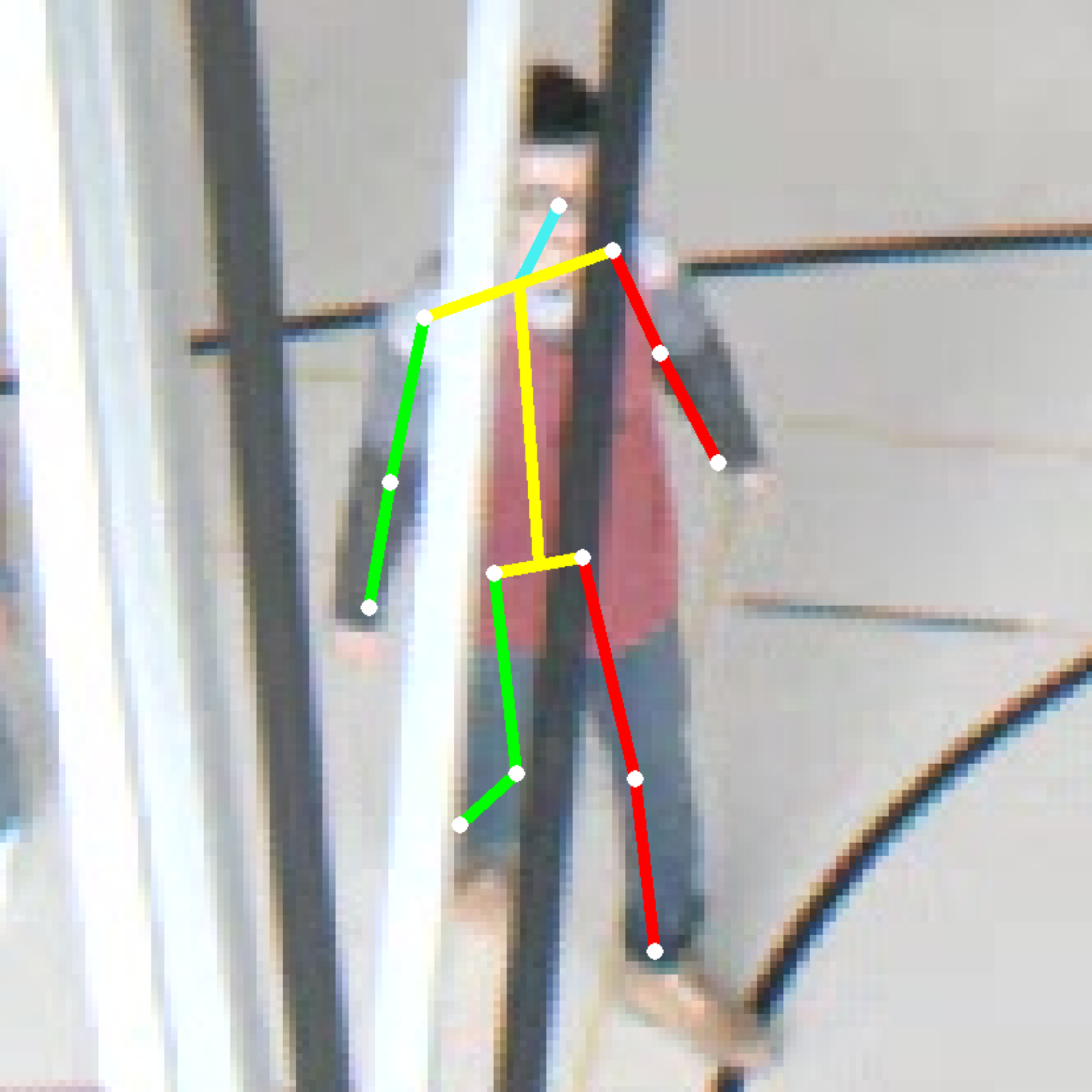}
    \end{subfigure}

    \begin{subfigure}[b]{0.3\linewidth}        
    	\centering
    	\includegraphics[width=\linewidth]{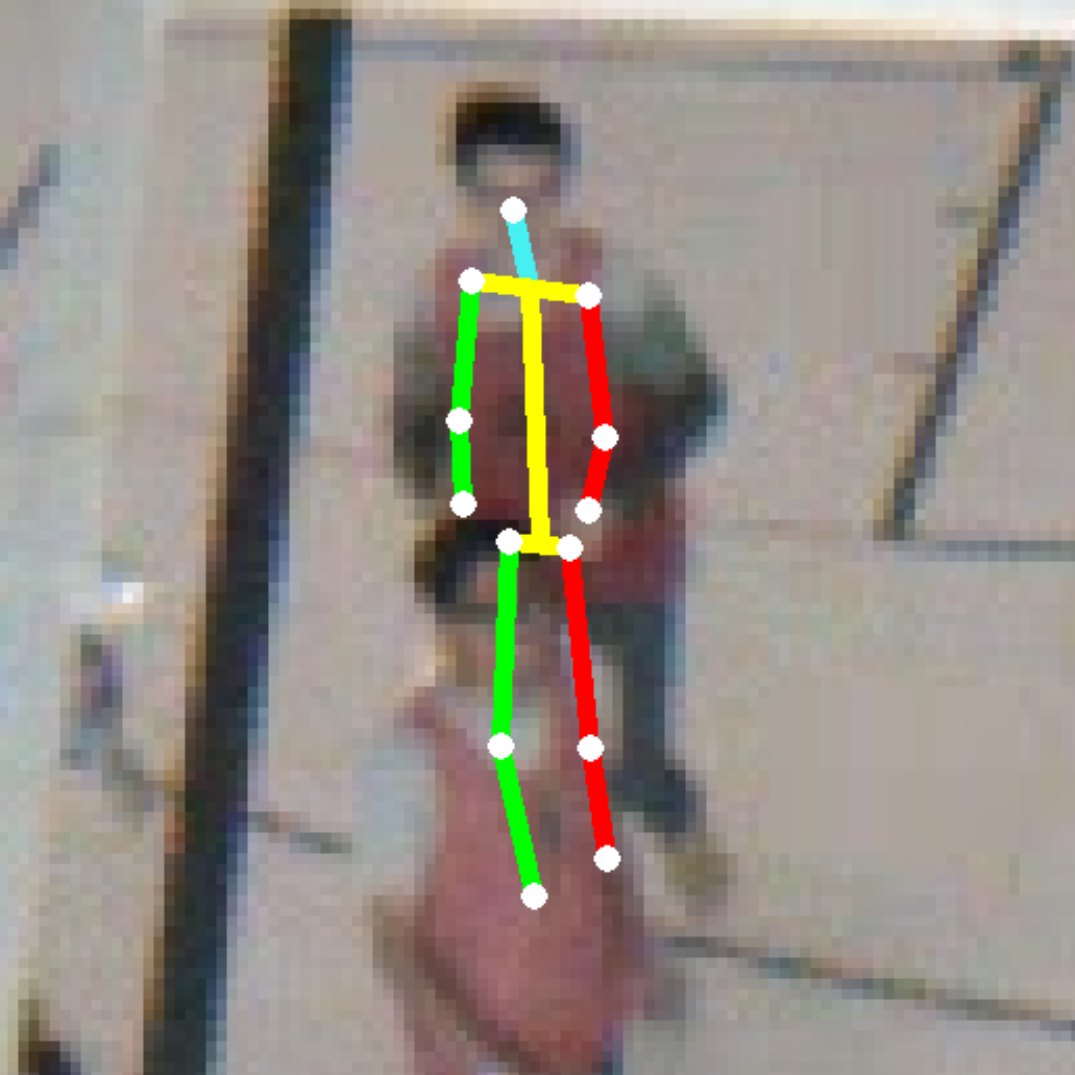}
     \caption{Iskakov~\etal~\cite{Iskakov19}}
    \end{subfigure}
    \begin{subfigure}[b]{0.3\linewidth}        
    	\centering
    	\includegraphics[width=\linewidth]{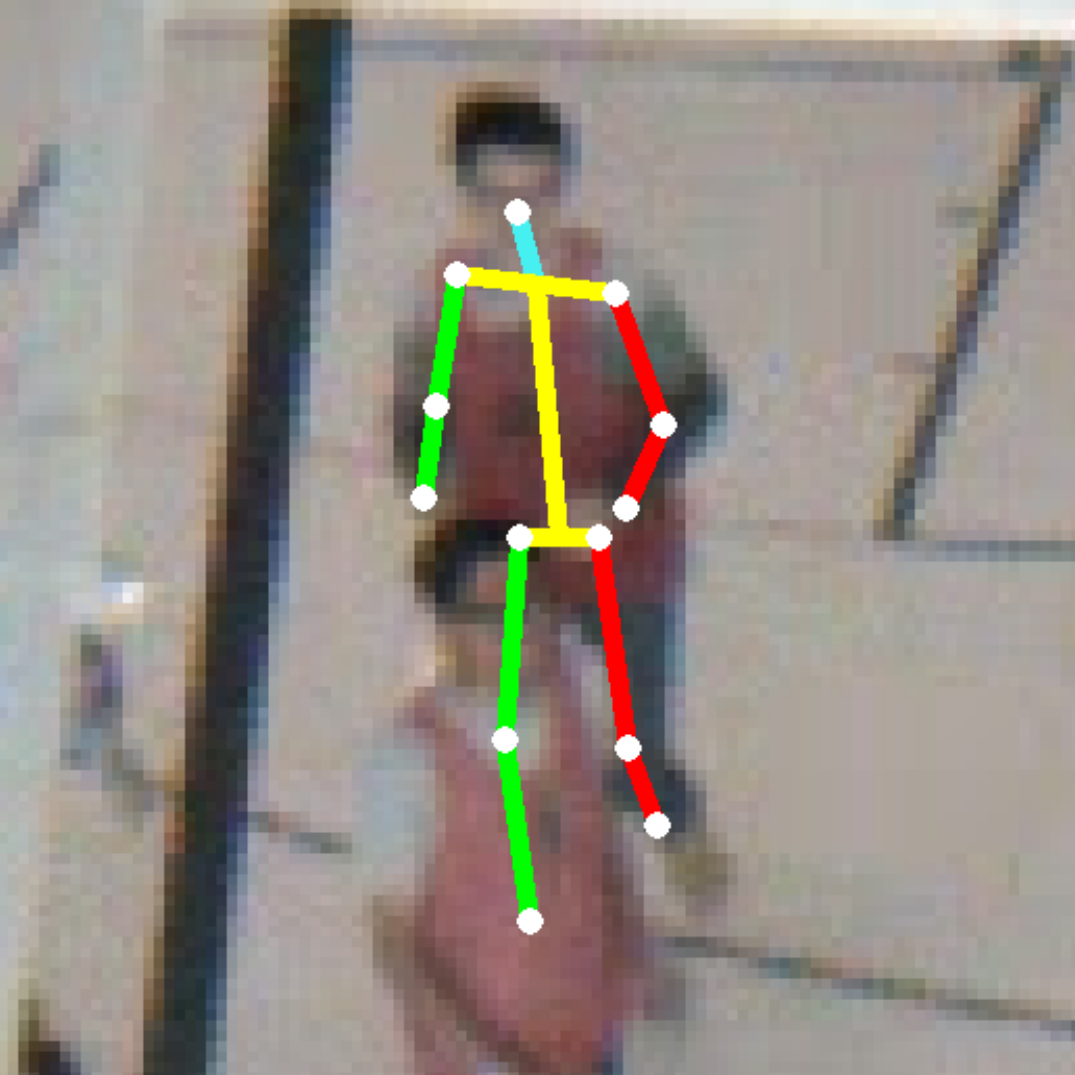}
            \caption{Roy~\etal~\cite{roy22a}}
    \end{subfigure}
    \begin{subfigure}[b]{0.3\linewidth}        
    	\centering
    	\includegraphics[width=\linewidth]{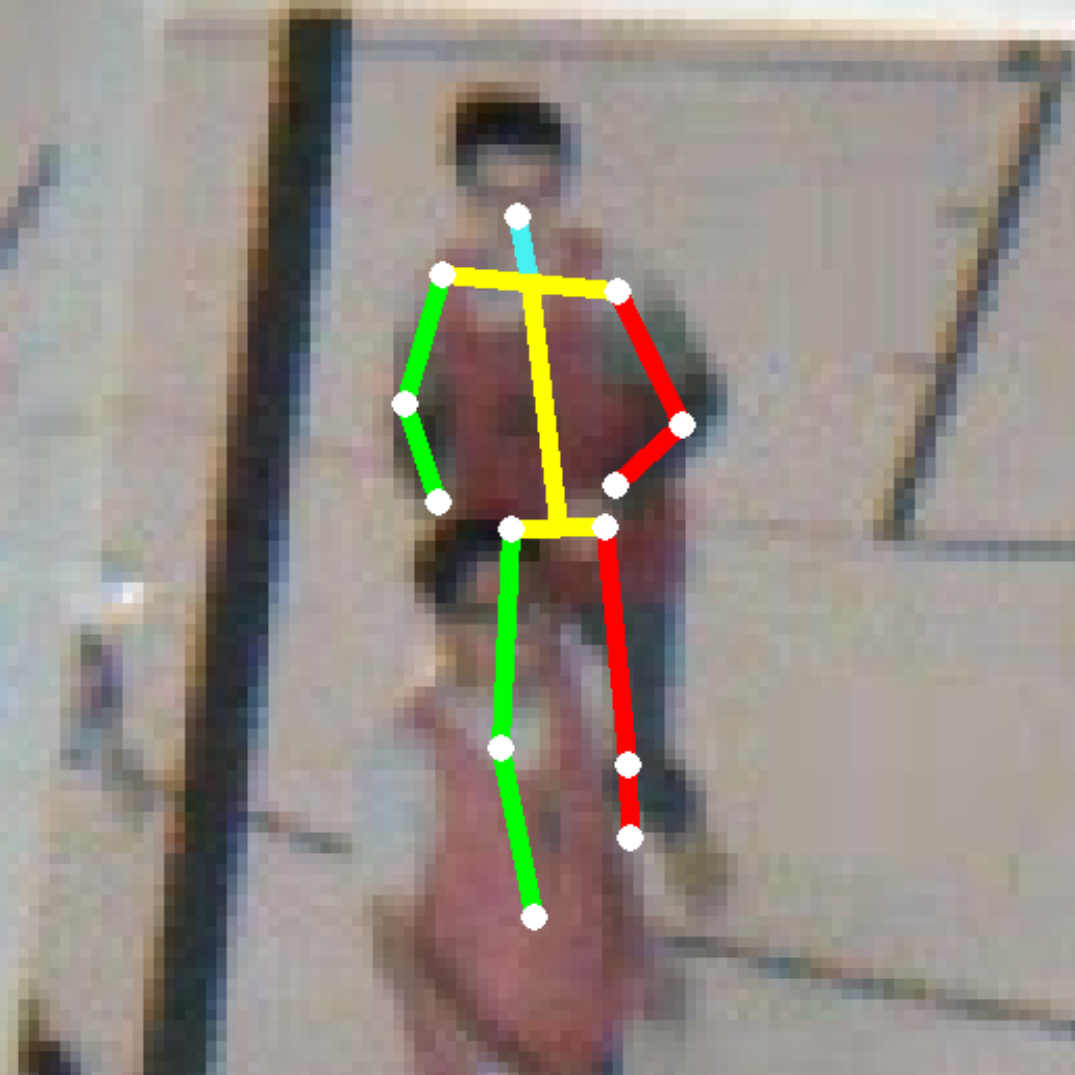}
            \caption{\textbf{Ours}}
    \end{subfigure}
    \caption{{\bf SportCenter.} Qualitative results on the samples from the ``Hard" test set for (a) Iskakov~\etal~\cite{Iskakov19}, (b) Roy~\etal~\cite{roy22a} and (c) Ours.}
    \label{fig:occlusion_images_sport_center_hard}
    \vspace{-0.25cm}
\end{figure}

\parag{SportCenter~\cite{roy22a}.} The images were captured during a game of amateur basketball featuring a total of 13 different people, with 10 of them being players at any given time. The images are captured using 8 fixed and calibrated cameras, out of which 6 cameras are used for pose estimation while the remaining 2 are used for tracking the players. The players are occluded either by other players or by various objects, such as the nets metal frames. Substantial changes in illumination occur, thus making pose-estimation even more challenging. The dataset consists of 315,000 images out of which only 560 images are provided with 3,740 2D annotated poses and 700 3D annotated poses. We use two subjects for evaluation, while the remaining ones are used for training in the semi-supervised setup. The test set is partioned into ``easy" and ``hard" cases, the difference being that the hard cases feature far stronger occlusions. Fig.~\ref{fig:occlusion_images_sport_center_hard} depicts four of them.

\parag{Metrics.} 

At training time, we learn weights for the whole network of Fig.~\ref{fig:overview} that is designed to operate on sequences. The result is a fully trained lifting network $LNet_{\Mat{\Phi}}$ that can operate on single views and that we test in this section. Hence, we report the Mean Per-Joint Position Error (MPJPE), the normalized NMPJPE, and the procrustes aligned PMPJPE in milimeters between the $LNet_{\Mat{\Phi}}$  predictions and the ground truth 3D poses. 


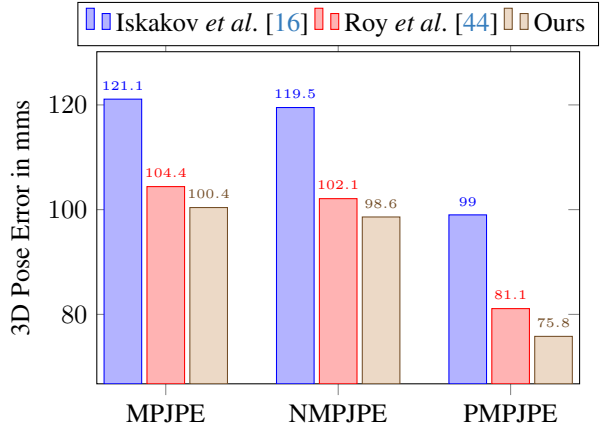
\begin{figure}[t]
    \centering

\begin{tikzpicture}  
\begin{axis}  
[  
    width=8cm,
    height=6cm,
    bar width=0.5cm,
    ybar, 
    enlargelimits=0.2,
    legend style={at={(0.5,1.15)}, 
      anchor=north,legend columns=-1},     
    ylabel={3D Pose Error in mms}, 
    symbolic x coords={MPJPE, NMPJPE, PMPJPE},  
    xtick=data,  
    nodes near coords,  
    every node near coord/.append style={font=\tiny},
    nodes near coords align={vertical},  
    ]  
\addplot coordinates {(MPJPE, 121.1) (NMPJPE, 119.5) (PMPJPE, 99)};
\addplot coordinates {(MPJPE, 104.4) (NMPJPE, 102.1) (PMPJPE, 81.1)};
\addplot coordinates {(MPJPE, 100.4) (NMPJPE, 98.6) (PMPJPE, 75.8)};
  \legend{Iskakov~\etal~\cite{Iskakov19}, Roy~\etal~\cite{roy22a}, Ours}
\end{axis}
\end{tikzpicture}  
    \caption{Comparative study on the SportCenter dataset in the semi-supervised setup.}
    \label{fig:res_sport_center}
\end{figure}

 
\begin{table}[t]
\centering
\caption{\small Quantitative results Human3.6M (Semi-supervised). Best results are shown in \textbf{bold}.}
	\scalebox{0.9}{
	\begin{tabular}{l|c|c|c}
		\toprule
		\multicolumn{4}{c}{10\% of All Data} \\
		\midrule
		\textbf{Method} & \textbf{MPJPE}~$\downarrow$ & \textbf{NMPJPE}~$\downarrow$ & \textbf{PMPJPE}~$\downarrow$ \\
		\midrule
		Kundu \textit{et.al.}~\cite{Kundu20} &  - & - & 50.8 \\
            Roy \etal~\cite{roy22a} & 56.9 & 56.6 & 45.4 \\
            \textbf{Ours} & \textbf{55.2} & \textbf{55.0} & \textbf{43.5} \\
		\bottomrule
		\multicolumn{4}{c}{Only S1} \\
		\toprule
		Rhodin \etal~\cite{Rhodin18b}  & 131.7 & 122.6 & 98.2 \\
		Pavlako \etal~\cite{Pavlakos19b}  & 110.7 & 97.6 & 74.5 \\
		Li  \etal~\cite{Li19c}  & 88.8 & 80.1 & 66.5 \\
		Rhodin  \etal~\cite{Rhodin18a}  & - & 80.1 & 65.1 \\
		Kocabas  \etal~\cite{Kocabas19}  & - & 67.0 & 60.2 \\
		Pavllo \etal~\cite{Pavllo19} & 64.7 & 61.8 & - \\
		Iqbal \etal~\cite{Iqbal20} & 62.8 & 59.6 & 51.4 \\
		Kundu  \etal~\cite{Kundu20} & - & - & 52 \\
            Roy \etal~\cite{roy22a} & 60.8 & 60.4 & 48.4 \\
		\midrule
            Ours  & \textbf{58.2} & \textbf{57.3} & \textbf{46.7} \\
		\bottomrule
	\end{tabular}
}	
\label{tab:h36m_semi} 
\end{table}

\subsection{Quantitative Results - Semi Supervised}

\paragraph{Human 3.6M.} 

We report the results for Human 3.6M data in Table~\ref{tab:h36m_semi}. Considering only 10\% of the dataset as the labeled set $\mL$, our approach consistently outperforms the best baseline~\cite{roy22a} in all three evaluation metric. We also outperform the other baselines when considering all the samples of subject S1 as the labeled set $\mL$ (Only S1).

This setup is more encouraging because some baselines~\cite{Kundu20, Iqbal20} rely on additional datasets and/or part based puppet models to instill prior human skeleton knowledge into  the underlying networks and to improve their generalization ability. In contrast, we use no such additional knowledge, which is advantageously replaced by the occlusion-resistant processing of our spatio-temporal graph. Another highlight is outperforming the approach in ~\cite{roy22a}, which models neither temporal consistency nor occlusions, showing the impact of these components.  

\begin{table}[t]
    \captionof{table}{{\small Quantitative results MPI-INF-3DHP (Semi-supervised). Best results are shown in \textbf{bold}.}}
\scalebox{0.92}{
	\begin{tabular}{l|c|c|c}
		\toprule		
		\textbf{Method} & \textbf{MPJPE}~$\downarrow$ & \textbf{NMPJPE}~$\downarrow$ & \textbf{PMPJPE}~$\downarrow$\\ 
		\midrule
		Rhodin  \etal~\cite{Rhodin18a}  & - & 121.8 & -   \\
		Kocabas  \etal~\cite{Kocabas19} & - & 119.9 & -  \\
		Iqbal \etal~\cite{Iqbal20}  & 113.8 & 102.2 & - \\
            Roy \etal~\cite{roy22a} & 102.2  & 99.6 & 93.6  \\
		\midrule 
		Ours & \textbf{100.1}  & \textbf{99.3} & \textbf{91.2} \\
		\bottomrule
	\end{tabular}
}	

\label{tab:mpii_semi}
\end{table}

\parag{MPI-INF-3DHP.} In Table~\ref{tab:mpii_semi}, we report similar results on the MPI-INF-3DHP dataset. As for H36M, we outperform the other competing methods by merely processing the masked graph $\mathcal{G}$ without adding any other prior knowledge.


\begin{table}[t]
\caption{Quantitative results on the ``Easy" and ``Hard" samples of the SportCenter dataset. We report the MPJPE in mms. The best results are \textbf{bold}. The baseline numbers are the published ones. }
\label{tab:sport_center_easy_hard}
\scalebox{.95}{
\begin{tabular}{|c|c|c|c|}
\toprule
Method  & Easy  & Hard  & All   \\
\midrule
Iskakov~\etal~\cite{Iskakov19} & 113.2 & 155.6 & 121.1 \\
Roy~\etal~\cite{roy22a}  & 98.7  & 139.2 & 104.4 \\
\midrule
Ours  &  \textbf{94.2} & \textbf{129.2} &  \textbf{100.4} \\
\bottomrule
\end{tabular}
}
\end{table}

\parag{SportCenter.} 

In  Fig.~\ref{fig:res_sport_center} and Table~\ref{tab:sport_center_easy_hard}, we compare our approach to the two methods~\cite{Iskakov19,roy22a} for which there are published results. Again we outperform them in all three evaluation metrics, especially on the hard samples that feature considerable amounts of occlusion. This confirms the importance of our proposed approach to masking joints during training. Fig.~\ref{fig:occlusion_images_sport_center_hard} provides a qualitative comparison between our method and the baselines~\cite{Iskakov19,roy22a} on ``Hard" samples from the test set. On these images, our method performs substantially better.  




\begin{table*}[t]
\centering
\caption{A comparative study of choosing different values of $\mathcal{S}$ that defines different temporal relationships between joints in~$\mathcal{G}$.}
\vspace{-.3cm}
\label{tab:temp_strides}
\scalebox{1}{
\begin{tabular}{|c|c|c|c|c|c|c|c|c|c|c|c|c|}
\hline
$\mathcal{S}$ & $\emptyset$ & $\left\{1\right\}$ & $\left\{3 \right\}$ & $\left\{5 \right\}$ & $\left\{7 \right\}$ & $\left\{1,3 \right\}$ & $\left\{1,5 \right\}$ & $\left\{1,7 \right\}$ &  $\left\{1,3,7 \right\}$ & $\left\{1,3,5,7 \right\}$ 
& $\left\{1,3,5,7,9,11 \right\}$\\ \midrule
\textbf{MPJPE}~$\downarrow$ & 62.1 & 61.0 & 60.7 & 60.5 & 60.4 & 59.9 & 59.3 & 59.1 & 58.6 & \textbf{58.2} 
& 58.3  \\ \hline
\end{tabular}
}
\vspace{-0.5cm}
\end{table*}
\subsection{Ablation Study}

In this section, we use the Human 3.6M~\cite{Ionescu14a} dataset to study the impact of varying the temporal parameters in the semi-supervised setup. We provide an additional ablation study about the mask generation parameters in the supplementary material.
Unless otherwise specified, the hyper-parameters are set to their default values, as discussed in Section~\textsection\ref{sec:implementation_details}.

\begin{table}[]
    \centering
    \caption{A comparative study of choosing the value of $t_p$ for the sequence length of T = 31 frames.}
    \vspace{-.3cm}
    \label{tab:choosing_tp}
    \scalebox{0.83}{
    \begin{tabular}{|c|c|c|c|c|c|c|c|c|}
    \toprule
$t_p$ & 0 & 2 & 4 & 8 & \begin{tabular}[c]{@{}c@{}}\textbf{15}\\ (\textbf{Ours})\end{tabular} & 20 & 25 & 30 \\
\midrule
MPJPE & 61.5 & 60.9 & 60.1 & 59.3 & \textbf{58.2} & 58.5 & 59.7 & 61.3 \\
    \bottomrule
    \end{tabular}
    }
\end{table}

\parag{Choosing the value of $t_p$.} 

Table~\ref{tab:choosing_tp} shows what happens when we choose different values for $t_p$,  the index of the target frame within a sequence of length $T=31$. The prediction accuracy drops significantly as we get closer to the end of the sequence, indicating that both passed and future poses contribute to prediction accuracy. It therefore makes sense to predict the pose at the center of the sequence, that is, $t_p=15$.


    
    

\begin{figure}
    \includegraphics[width=1.1\textwidth,keepaspectratio]{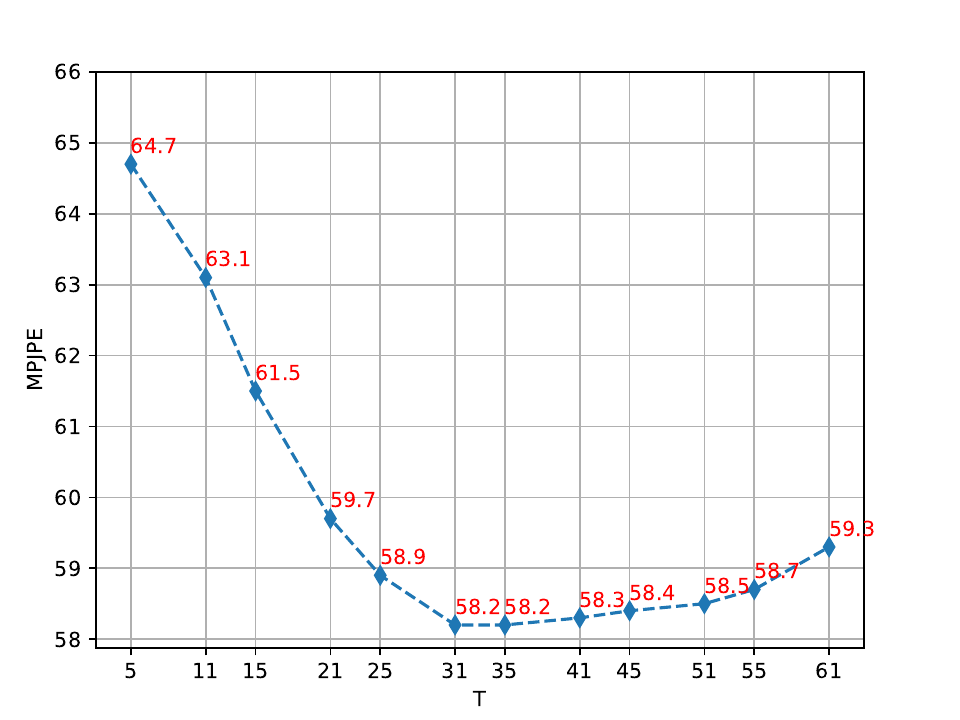}
    \caption{A comparative study choosing different values of sequence length T.}
    \label{fig:window_size}
    \vspace{-.5cm}
\end{figure}

\parag{Choosing the sequence length $\textrm{T}$.} 
Fig.~\ref{fig:window_size} illustrates what happens for different values of the sequence length $\textrm{T}$ from 5 to 61. For smaller values of $\textrm{T}$, the human body does not undergo much deformation. As a result, the graph $\mathcal{G}$ does not contain much temporal information, which results in higher prediction errors. As we increase $\textrm{T}$, we get the smallest error for $\textrm{T}=31$. After that, the error increases again because the poses within the sequence become too diverse.

\parag{Choosing different temporal strides $\Delta$ in ~$\mathcal{S}$.}

Table~\ref{tab:temp_strides} shows the consequences of including different temporal strides in the set $\mathcal{S}$. $\mathcal{S}=\emptyset$ yields a graph $\mathcal{G}$ containing only spatial connections, which unsurprisingly yields higher errors by ignoring temporal consistency. Successive additions of temporal connections progressively boosts performance as the refinement network can then exploit diverse, yet complementary temporal relationships between the nodes in $\mathcal{G}$. The best results are obtained for $\mathcal{S}=\left\{1, 3, 5, 7 \right\}$. Adding even more temporal edges does not yield any additional improvements as these extra temporal edges do not model any new inter-joint temporal dependencies.

\begin{figure}
    \centering
    \includegraphics[trim={0 0 0 1cm},clip, width=1.1\textwidth,keepaspectratio]{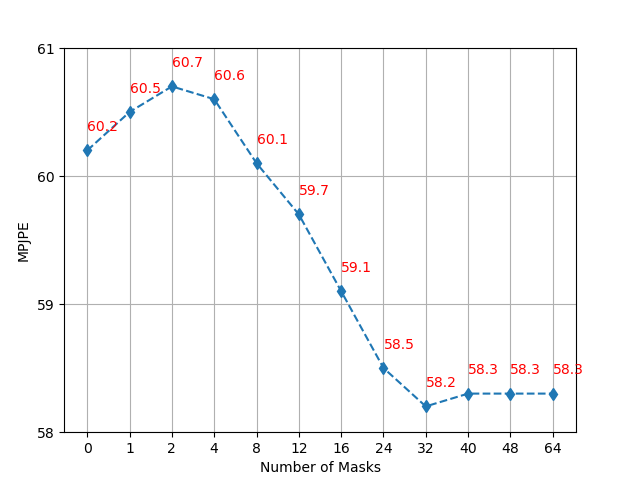}
    \caption{A comparative study of generating different number of structured masks using~\cite{Durasov21a}}
    \label{fig:num_masks}
    \vspace{-.3cm}
\end{figure}

\begin{table}[ht]
    \centering
    \caption{Dropping nodes in $\mathcal{G}$ using structured masks (Ours) vs MC-Dropout~\cite{Gal16a}. $\tau$ denotes the dropout rate.}
    \label{tab:structured_vs_random}
    \vspace{-.3cm}
    \scalebox{0.9}{
    \begin{tabular}{c|cccccc|c}
\cline{2-7}
 & \multicolumn{6}{c|}{MC-Dropout~\cite{Gal16a}} &  \\ \hline
 \multicolumn{1}{|c|}{$\tau$} & \multicolumn{1}{c|}{0.0} & \multicolumn{1}{c|}{0.1} & \multicolumn{1}{c|}{0.2} & \multicolumn{1}{c|}{0.3} & \multicolumn{1}{c|}{0.4} & 0.5 & \multicolumn{1}{c|}{\textbf{Ours}} \\ \hline

  \multicolumn{1}{|c|}{\textbf{MPJPE}~$\downarrow$} & \multicolumn{1}{c|}{60.2} & \multicolumn{1}{c|}{60.9} & \multicolumn{1}{c|}{61.5} & \multicolumn{1}{c|}{61.8} & \multicolumn{1}{c|}{62.4} & 63.1 & \multicolumn{1}{c|}{\textbf{58.2}} \\ \hline
\end{tabular}
}
\end{table}

Here, we study the impact of different parameters, \ie~ $\textrm{N}_{\textrm{M}},~\beta~\textrm{and}~\alpha$, of the mask generation process$~\Omega$ in the overall performance of our proposed method (Refer to Section~\textsection~\textcolor{red}{3.3.1} in the main text for more details) on the Human 3.6M dataset~\cite{Ionescu14a} in the semi-supervised setup.


\subsubsection*{Structured Masks~\cite{Durasov21a} vs MC-Dropout~\cite{Gal16a}.} 

Recall that we use structured masks to drop nodes from the spatio-temporal graph  $\mathcal{G}$ to produce  $\bar{\mathcal{G}}$, which is then fed to the refinement network.  Table \ref{tab:structured_vs_random} shows the impact of randomly dropping graph nodes,  as in MC-Dropout~\cite{Gal16a}, with a dropout rate of $\tau$.  Hence, $\tau=0.0$ means that no nodes are dropped. We do better for all values of $\tau$. 

\begin{table}[]
    \centering
    \caption{Varying the parameter $\alpha$ that controls the amount of overlap between masks. $\bar{\tau}$ denotes the rate of nodes masked in~$\mathcal{G}$. It plays the same role as the dropout rate $\tau$ for MC-Dropout~\cite{Gal16a}.
    }
    \vspace{-.3cm}
    \label{tab:alpha}
    \scalebox{0.8}{
    \begin{tabular}{|c|c|c|c|c|c|c|c|c|}
    \hline
    $\alpha$ & 1.0 & 1.2 & 1.4 & 1.8 & 2.0 & 2.5 & 3.0 & 3.5 \\ 
    \hline
    $\bar{\tau}$ & 0 & 0.17 & 0.25 & 0.33 & 0.5 & 0.60 & 0.67 & 0.72 \\ 
    \hline
    \textbf{MPJPE}~$\downarrow$ & 60.2 & 60.1 & 59.6 & \textbf{58.2} & 59 & 61.2 & 63.6 & 65.4 \\ \hline
    \end{tabular}
    }
    
    \vspace{-0.5cm}
\end{table}
\vspace{-0.5cm}
\subsubsection*{Mask Overlap ($\alpha$ parameter)}

In Table \ref{tab:alpha}, we report the impact of varying the $\alpha$, the parameter that controls the amount of overlap between the generated masks $\Mat{M}_i \in \mathcal{M}$. It also decides the number of ones (\ie~$\beta$) in each of the masks $\Mat{M}_i$
. All the nodes are present in the graph~$\bar{\mathcal{G}}$ when $\alpha$ is set to 1. As we increase the value of $\alpha$, more nodes are structurally dropped from the graph $\mathcal{G}$ to obtain~$\bar{\mathcal{G}}$, with the lowest error obtained when $\alpha=1.8$. As we further increase $\alpha$, more nodes are dropped from $\mathcal{G}$. This results in not having sufficient number of nodes in~$\bar{\mathcal{G}}$ that are needed to preserve the spatial and temporal dependencies between the nodes, leading to higher error.


\subsubsection*{Number of parameters ($\textrm{N}_{\textrm{M}}$ parameter)} 

Fig.~\ref{fig:num_masks} shows the performance of our approach as a function of $\textrm{N}_{\textrm{M}}$, the number of masks in~$\mathcal{M}$. For $\textrm{N}_{\textrm{M}} = 0$, we do not use any mask over the graph~$\mathcal{G}$, thereby indicating that none of the nodes are masked in~$\mathcal{G}$ to obtain~$\bar{\mathcal{G}}$. Initially the performance degrades as we increase the number of masks $\textrm{N}_{\textrm{M}}$. However we see a boost in the performance from $\textrm{N}_{\textrm{M}}=8$ as the generated masks become more diverse with the best result obtained for $\textrm{N}_{\textrm{M}}=32$. This in turn aids our networks to efficiently model different forms of occlusions. The error plateaus after that as the masks are not diverse enough to capture any new occlusion pattern.

\begin{table}[t]
    \centering
    \caption{Quantitative results Human3.6M (Fully-supervised).
    }
    \label{tab:h36m_fully}
		\scalebox{0.85}{
			\begin{tabular}{|l|c|c|c|}
				\toprule
				\textbf{Methods} &  \textbf{MPJPE}~$\downarrow$ & \textbf{NMPJE}~$\downarrow$ & \textbf{PMPJE}~$\downarrow$ \\
            \midrule
            		Rogez  \etal~\cite{Rogez17}  & 87.7 & - & 71.6 \\
				Rhodin \etal~\cite{Rhodin18a}  & 66.8 & - & - \\
				Zhou   \etal~\cite{Zhou17f} & 64.9 & - & - \\
				Martinez \etal~\cite{Martinez17a}  & 62.9 & - & 52.1 \\
				Sun~\etal~\cite{Sun17}  & 59.6 & - & - \\
				Yang \etal~\cite{Yang18b} & 58.6 & - & - \\
				Pavlakos \etal~\cite{Pavlakos18a} & 56.2 & - & - \\
				
				Habibie \etal~\cite{Habibie19}  & - & 65.7 & - \\
				Kocabas \etal~\cite{Kocabas19} & 51.8 & 51.6 & 45.0 \\
								Iqbal \etal~\cite{Iqbal20} & 50.2 &  \textbf{49.9} & \textbf{36.9} \\
				Kundu \etal~\cite{Kundu20} & - & - & 50 \\
				Roy~\etal~\cite{roy22a} & 55.3 & 52.4 & 44.2 \\
				Ci~\etal~\cite{Ci19} & 52.7 & - & - \\
				Pavllo~\etal~\cite{Pavllo19} & 51.8 & - &  40.0 \\
				Liu~\etal~\cite{Liu20f} & 52.4 & - & - \\
                    Iskakov~\etal~\cite{Iskakov19} (vol.) & 49.9 & - & - \\
                    Cai~\etal~\cite{Cai19} & 50.6 & - & - \\
                    Zeng~\etal~\cite{Zeng20} & 49.9 & - & - \\
				Zou~\etal~\cite{Zou21} & \textbf{49.4} & - & 39.1 \\
                    Xu~\etal~\cite{Xu21b} & 51.9 & - & - \\
                    
                    \midrule\
				Ours  & 51.0 & 50.2 & 39.4 \\
                    \bottomrule
			\end{tabular}
		}
    
\end{table}

\section{Quantitative Results - Fully Supervised} 
\label{sec:fullResults}

We now evaluate our approach on the Human 3.6M and MPI-INF-3DHP datasets in the fully supervised setup: We use {\it all} the  2D and 3D ground truth training annotations that are provided. We report the results for H36M and MPI in Table~\ref{tab:h36m_fully} and~\ref{tab:mpi_fully}. To generate the numbers for~\cite{roy22a}, we used our own implementation. The others are the published ones.

Here, the result are more mixed. Our approach does well overall but the methods of~\cite{Cai19,Iqbal20,Zeng20,Zou21,Chen21} outperform us on some of measures in one of the two datasets. We attribute this to the following differences: First, a pipeline such as that of~\cite{Zou21} relies on a Cascaded Pyramid Network 2D point detector~\cite{Chen18e}, which is reported to be more powerful that the one we use~\cite{roy22a}. We therefore tried using it but this did not change the overall performance of our approach. This suggests that we need to use a more complex deeper backbone network similar to the one used in~\cite{Zou21} to take advantage of it. Second, the method of \cite{Iqbal20} leverages additional datasets in training of their estimation networks, while the approaches of~\cite{Zou21,Cai19,Zeng20} enforce spatial constraints on the weights learned by the GCNN based estimation networks. We do neither of these things, but they could be integrated into our own GCNNs. In short, even though our network operates under more challenging conditions, it still delivers comparable results, which confirms the power of our occlusion-handling scheme. Moreover, we do not enforce any additional weight modulation and sharing practices of~\cite{Zou21,Cai19,Zeng20}, which are key for their good performance in the fully supervised setup. Furthermore, the methods of~\cite{Cai19,Zeng20,Zou21,Chen21} have not been demonstrated in the semi-supervised framework and it is not entirely clear what they would do in this training mode. 


\begin{table}[t]
    \centering
    \caption{Quantitative results MPI-INF-3DHP (Fully-supervised). Best results are shown in \textbf{bold}.$\dagger$ indicates cross dataset evaluation without further fine-tuning.}
    \label{tab:mpi_fully}
    \scalebox{0.95}{
    \begin{tabular}{|l|c|c|}
    \toprule
Method & \textbf{3DPCK$\uparrow$} & \textbf{MPJPE$\downarrow$} \\
\midrule
Mehta~\etal~\cite{Mehta17a} & 75.7 & 117.6 \\
Mehta~\etal~\cite{Mehta17b} & 76.6 & 124.7 \\
Pavllo~\etal~\cite{Pavllo19} & 86.0  & 84.0 \\
Chen~\etal~\cite{Chen21} & \textbf{87.9} & 78.8  \\
Zeng~\etal~\cite{Zeng20}$\dagger$ & 77.6 & - \\
Ci~\etal~\cite{Ci19}$\dagger$ & 74.0 & - \\
Xu~\etal~\cite{Xu21b}~$\dagger$ & 80.1 & - \\
Zou~\etal~\cite{Zou21} &  86.1 & - \\
 \midrule
Ours & 85.6 & \textbf{78.1} \\
\bottomrule
\end{tabular}
}
\end{table}

    \begin{table}[]
\caption{A comparative study of the performance of our proposed method using the 2D detections of Roy~\etal~\cite{roy22a} vs 2D Mocap key-points on the Human 3.6M dataset.}
\label{tab:h36m_2d_gt}
\vspace{-.3cm}
\begin{tabular}{|c|ccc|}
\toprule

Method & \textbf{MPJPE}~$\downarrow$ & \textbf{NMPJPE}~$\downarrow$ & \textbf{PMPJPE} $\downarrow$ \\


\midrule
Roy~\etal ~\cite{roy22a}& 51.0 & 50.2 & 39.4 \\ 
Mocap & 42.5 & 43.0 & 33.8  \\

 \bottomrule
\end{tabular}
\vspace{-0.5cm}
\end{table}

\section{Limitations}
A limitation of our proposed approach is that it relies on a pre-trained joint 2D detector~\cite{roy22a}, whose effectiveness or lack thereof directly impacts our final results. To demonstrate this, we ran an experiment in which we replaced the 2D detections by their ground-truth values and report the results in Table~\ref{tab:h36m_2d_gt}. Unsurprisingly, the metrics are substantially improved. This points toward the fact that the 2D detector should be trained as well and made part of the trainable end-to-end pipeline.   


\section{Conclusion}

We have proposed an effective approach to enhancing the robustness to occlusions of a single-view 3D human pose estimation network. To this end, we perform graph convolutions over a spatio-temporal graph defined on the 3D joints predicted by a lifting network. During training, we use structured binary masks to disable some nodes together with the corresponding edges. This mimics the fact some joints can be hidden over a given period of time and trains our lifting network to be resilient to that. 

Our experiments show that this simple method delivers excellent results and enables us to outperform more complex state-of-the-art methods on semi-supervised scenarios. However, they also show that some of the priors that these methods exploit could be profitably incorporated into our framework for a further performance boost. We will focus on this in future work.


    

{
    \small
    \bibliographystyle{ieeenat_fullname}
    \bibliography{bib/string,bib/vision,bib/learning,bib/photog,bib/optim,bib/soumava_refs}
}

\end{document}